\documentclass[review]{elsarticle}
\usepackage{hyperref}

\journal{Medical Image Analysis}

\begin{document}
\begin{frontmatter}

\title{Automatic 3D modelling of craniofacial form}

\author{Nick Pears}
\address{Department of Computer Science, University of York, York, UK. }

\author{Christian Duncan}
\address{Craniofacial Unit, Alder Hey hospital, Liverpool, UK}

\begin{abstract}
Three-dimensional models of craniofacial variation over the general population are useful for assessing pre- and post-operative head shape
when treating various craniofacial conditions, such as craniosynostosis. We present a new method of automatically building both sagittal profile models and full 3D surface models of the human head using a range of techniques in 3D surface image analysis; in particular, automatic facial landmarking using supervised machine learning, global and local symmetry plane detection using a variant of trimmed iterative closest points, locally-affine template warping (for full 3D models) and a novel pose normalisation using robust iterative ellipse fitting. The PCA-based models built using the new pose normalisation are more compact than those using Generalised Procrustes Analysis and we demonstrate their utility in a clinical case study.
\end{abstract}

\begin{keyword}
3D shape modelling \sep Symmetry plane extraction \sep Automatic landmarking  \sep 3D feature matching
\end{keyword}

\end{frontmatter}


\section{Introduction}

Modern techniques in 3D surface imaging are now making very high resolution 3D surface images available to clinicians and other medical professionals. However, the availability of tools for 3D surface analysis is limited and there is an urgent need for reliable methods
of automatic 3D surface analysis and automatic 3D surface modelling tools. Such tools can generate quantitative analyses for both surgical planning and surgical outcome assessment in craniofacial surgery, where currently assessments are of a qualitative nature.

This paper is concerned with the automatic 3D modelling of human head form (shape and size) using the \emph{Headspace} dataset and its application in a clinical context. Although there are existing 3D models of the human face, perhaps the most well-known of which is the \emph{Basel Face Model} \cite{bfm09}, to our knowledge, there is no existing 3D model of the full human head that includes both face \emph{and} full cranium and which can accurately and automatically parameterise the 3D form of a wide range of children's and adult's faces. 
This is challenging due to diverse facial forms of the general population engendered by the two sexes and the general population's wide ranges of age and ethnicity. 

Here we present our automatic modelling process that build statistical 3D shape models that will form the core of new model-driven software tools. Such tools, for example, will be able to indicate a target transformation of a dysmorphic (eg. Apert syndrome) face and head shape 
as a mapping from the current shape into a modelled face space region labelled \emph{non-dysmorphic}.

In this paper, we first build a model of sagittal head profile variation, where the profile is extracted using multiple planes of local bilateral symmetry. Outputs from this 2D model build are then employed in the pose normalisation required for a full 3D surface model build.
As this is our first model, we initially focus on male head shape, using a subset of 100 adult male 3D head scans. We plan to scale this up and analyse head shape of females and the training population as a whole in the near future. The techniques used in terms of machine learning are well-established and standard eg. Gaussian modelling of distributions, landmark scoring functions based on Linear Discriminant Analysis (LDA) and linear shape modelling based on Principal Component Analysis (PCA). The modelling processes have been successful because they are well-constrained by geometric processes; for example those that consider local symmetry of the face and the elliptical shape of the cranium.

As well as these being the first statistical models that include the full cranium of the human head, the model building processes are fully automatic (after a supervised training phase) and can deal with the head in arbitrary initial pose. Further contributions include a novel operator to refine landmark positions and a novel pose normalisation scheme based on robust ellipse fitting. This normalisation provides more compact statistical models that the standard pose normalisation generated by Generalised Procrustes Analysis (GPA).
Outputs from the model building process have allowed us to make a number of interesting observations on head shape over the training set, such as common asymmetries in the nasal area, regularities in the orientation of the fitted ellipse and the principal modes of head shape variation.

The remainder of this paper is structured as follows: in Sect. \ref{sec:litReview}, we discuss related literature and in the following section, we describe the primary dataset used to train the model, known as the \emph{Headspace} dataset. In Sect. \ref{sec:overview}.
we give an overview of how our automatic modelling system works. This is followed by a section giving descriptions of each process in more detail, with intermediate outputs where appropriate. In Sect. \ref{sec:evaluation}, we evaluate our models in terms of compactness for different pose normalisation schemes and when using both automatic and manual facial profile landmarks. 
In Sect. \ref{sec:3Dmodelling}, we show how our cranial profile model can be extended to a full 3D surface model, and we build such a model.
In Sect. \ref{sec:caseStudy}, the final section before conclusions, we present a clinical case study of 2D sagittal profile model use.

\section{Related literature}
\label{sec:litReview}

Relevant topics in the literature include 3D shape registration, 3D shape modelling and 3D model fitting. Additionally, there are many similar techniques in the literature that have been applied to 2D image applications and given that a 3D image (or 3D scan) usually includes a standard 2D colour-texture image channel as well as a shape channel, 2D image registration is also relevant. Furthermore, there are similar model building and registration themes in many research fields, such as medical imaging, computer vision, computer graphics, anthropometry, orthodontics, and in the morphometrics literature associated with zoology, anthropology and palaeontology. Thus previous work relevant to computational 3D shape modelling is vast and, here, we can only give a few key milestones relevant to automatic 3D modelling. 

Early work in organic shape analysis was performed on 2D images. Thompson's \emph{On Growth and Form} (1917) \cite{Thompson1992} sketched transformation grids to show how one shape had to be deformed to match another in terms of their corresponding landmarks. 
In the 1980s, statistical developments concerning the shape space for Procrustes-registered data derived from the seminal work of Kendall \cite{Kendall1984}. Bookstein and his work on thin plate splines \cite{Bookstein1989} was central to the development of the related field of statistical shape analysis \cite{DrydenMardia1998}, which provides the theoretical underpinnings of the statistical modelling of organic form. Later Bookstein published work that employed a combination of Procrustes analysis and TPS in order to analyse shapes in terms of their landmark and semi-landmark positions \cite{Bookstein1996}.

Also in the 1980s, relevant work was produced in the fields of Robotics and Computer Vision. 
For example, \emph{rigid transformation estimation} between a pair of 3D shapes was solved by several 
Computer Vision researchers including  Faugeras, Horn, and Arun. 
Approaches based on Singular Value Decomposition (SVD) that lead to least squares solutions are particularly popular \cite{Arun1987}.

Statistical modelling of the 2D (texture) variance of faces was also introduced in the 1980s, with Sirovich and Kirby applying PCA to facial images. In the early 90s, two groups of researchers (Besl and McKay, and Chen and Medioni) independently proposed the \emph{Iterative Closest Points} (ICP) algorithm, which cycles through three main steps: i) finding surface correspondences as closest points; ii) computing the rigid transformation estimation between them, and iii) applying the transformation to one of the surfaces to make it closer to the other.  The pioneering approaches of Besl and McKay \cite{Besl92} and Chen and Medioni \cite{Chen92} differ in the way that they compute the `closest points'. Later several research groups investigated non-rigid registration approaches where, for example, local affine deformations are permitted \cite{Allen2003} \cite{Amberg07}.

In the 1990s, a Manchester team developed shape models applied to 2D images, termed
\emph{Point Distribution Models} (PDM). The work is done with reference to 2D shapes, 
where corresponding points are manually marked on the boundaries of a set of training examples.
The points are aligned to minimise the variance in distance between corresponding points. This is done 
by encapsulating a Procrustes-based alignment in an iterative procedure, where the mean is normalised to a default scale and pose at each iteration. After such alignment, a standard principal component analysis (PCA) captures how the shapes deviate from the mean shape. 

Model fitting is also presented in \cite{Cootes1995}, where pose, scale and shape parameters are determined in order to fit the model to an image.
The authors termed the work \emph{Active Shape Models}, perhaps to link it with earlier work
on Active Contour Models (``Snakes'') \cite{Kass88}.The same research team also went on to include texture in their models to give \emph{active appearance models} \cite{Cootes2001}.
The same Manchester team developed a set of shape modelling approaches where the best correspondences are those that define the most compact shape model given some quality of fit between the model and the data  \cite{Davies2001} \cite{Kotcheff1998}. 

The meshes that build the model must be approximately aligned and, in some applications, the only way to do this robustly is to detect and label features (aligning principal axes of scan data is often not sufficiently accurate). Meshes are then reparameterised to some form of $(u,v)$ space, such that correspondences have the same parameterisation: i.e. the same $(u,v)$ coordinates across $N$ training scans defines the correspondences. The individual training shape parameterisations are then adjusted by applying smooth mappings in the parameter space and the cost function is minimised. If a global minimum of their objective function is found, then the optimal \emph{groupwise} 
(i.e. across all training scans) correspondences give the most compact shape model. Kotcheff and Taylor \cite{Kotcheff1998}, defined the \emph{DetCov} metric, which minimises the determinant of the covariance matrix associated with the training data correspondences. 
A genetic algorithm was used to perform the minimisation, although this has the potential of being slow to converge.
Later, Davies et al used an optimisation to achieve \emph{Minimum Description Length} \cite{Davies2001}. 
This objective function captures both the model complexity itself and the data values required to express each of the training scans with the model. This objective function can trade off complexity of the model and quality of fit.

In the late 1990s, Blanz and Vetter built a `3D morphable model' (3DMM) from 3D face scans
and published their seminal SIGGRAPH paper  \cite{Blanz1999}. 
and employed it in a 2D face recognition application \cite{Blanz2003}.
in an `analysis by synthesis' approach.
that aimed to tackle recognition difficulties associated with varying illumination and head pose. 
The key point is that 
The use of a 3D face model allows the separation of shape from pose, and albedo from illumination.
In the original Blanz-Vetter model, 
two hundred 70,000 vertex scans were employed (young adults, 100 males and 100 females). 
Dense correspondences were computed using a gradient-based optic flow algorithm - both shape and colour-texture is used. 
This is unreliable in smooth areas and so a smooth interpolation of flow vectors is performed and both optic flow and smoothing are performed iteratively through coarse-to-fine levels of resolution. 
The model is constructed by applying PCA to shape and colour-texture (separately). 
A fitting process is described where the morphable model is matched to 2D images, and hence models of illumination and imaging are employed. 
The fitting process, which uses stochastic gradient descent, adjusts parameters of the morphable model, illumination and imaging, in order to minimise the Euclidean distance error between a synthetic generated image and the image being fitted. 

The \emph{Basel Face Model} (BFM) was developed by Paysan et al \cite{bfm09} at the University of Basel. 
Here the method of determining corresponding points was improved (cf. Blanz and Vetter's pioneering work). 
Instead of optic flow, a set of hand-labelled feature points is marked on each of the 200 training scans. 
The corresponding points are known on a template mesh, which is then morphed onto the training scan using local affine transformations \cite{Amberg07}. This work was inspired by \cite{Allen2003} who first proposed using underconstrained per-vertex affine transformations, where additional constraints are enforced by limiting the movement between neighbouring points in the template mesh. Allen et al refer to the regularisation as minimising a smoothness error, whereas Amberg et al refer to the regularisation as a stiffness term. There are further non-rigid point set registration approaches in the literature around the same time as Amberg et al's work, such as the Levenberg-Marquardt non-linear optimisation approach of Li et al \cite{Li2008} and the Expectation-Maximisation approach of Myronenko and Song \cite{Myronenko2010}.

In statistical model building, groupwise optimisation of registration can offer better performance than a sequence of pairwise registrations (although pairwise registration can be used to initialise a groupwise registration). However, the cost of a groupwise registration is that the optimal solution resides in a much higher dimensional search space.

Sidorov et al. \cite{Sidorov2009} presented groupwise non-rigid (2D) image registration using an optimisation approach called Simultaneous Pertubation Stochastic Approximation (SPSA). This optimisation does not become prohibitively expensive over a large number of cost function parameters. The parameters are those defining the pixelwise deformation field of image displacements required to incrementally improve registration. For fast GPU implementation, the cost function is the average of pixelwise discrepancy between each of the N warped images and the average of the remaining N-1 warped images.
Later, the same authors extended this to textured 3D surfaces, where surfaces were first isometrically embedded in a 2D plane using Multi Dimensional Scaling (MDS) operating on pairwise geodesic distances \cite{Sidorov2011}. 
Geodesic distances were extracted using a fast marching method. 
Given an initial affine registration of the embedded images, SPSA was employed to optimise the groupwise registration. Thus 3D registration is achieved entirely on the basis of 2D image data, although the method could be extended to 3D local shape descriptors. 
The system was able to register 3D scans of 32 different subjects, but performance of the resulting shape model was not detailed. 

Recent work at York \cite{Creusot2013}, gives a highly robust approach to sparse model fitting, based on random sample consensus (RANSAC) and learning `per landmark' detector functions using LDA.
A landmarking algorithm that uses the same overall structure is employed in this work. 
Further work on landmark saliency shows how to automatically learn the landmarks that constitute a sparse face model from a set of registered 3D face scans \cite{Creusot2012}.

\section{Overview of the headspace dataset}
\label{sec:headspace}

This work used the \emph{Headspace} dataset, which was collected by Alder Hey hospital craniofacial unit (Liverpool, UK).
This unit performs surgery for patients with craniofacial conditions that impact both neurodevelopment and appearance. 
The aim of surgery is to protect vital structures (brain, eyes, airway) and normalise appearance such that patients obtain maximal quality of life through optimal functioning and avoidance of stigmatisation because of their conditions.  

\begin{center}
\begin{figure}
\begin{tabular}{cc}
\includegraphics[width=0.40\linewidth]{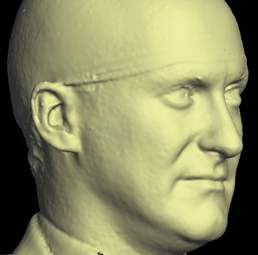} & \includegraphics[width=0.38\linewidth]{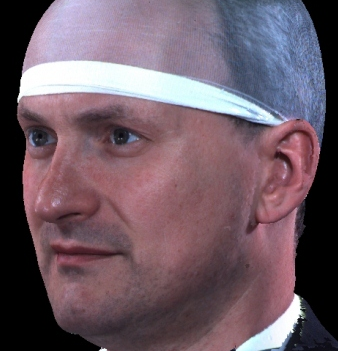} \\
\end{tabular}
\caption{The 3D data channel (left) and 3D+2D channels with pasted on texture (right)}
\label{fig:Christian}
\end{figure}
\end{center}

\begin{center}
\begin{figure}
\includegraphics[width=1.00\linewidth]{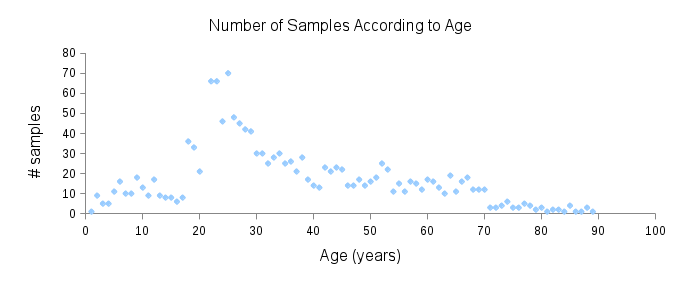}
\caption{Distribution of age of imaged subjects in the Headspace dataset}
\label{fig:headspaceAge}
\end{figure}
\end{center}

Toolsets to define normal appearance and therefore both assist with planning of surgery and  define optimal outcomes from a range of surgical interventions are currently either crude or lacking.  Therefore, the unit collected a large, high quality 3D image dataset of 1523 human heads. An example is given in Fig. \ref{fig:Christian} showing the 3D channel only on the left, and the 3D data with texture pasted on on the right. 
The data was collected over well-conserved demographics, and an age range of 1 to 89 years, see Fig. \ref{fig:headspaceAge}. The dataset was collected with the aim of developing a normal equivalent of the human cranium and face in order develop the tools to enhance patient experience following surgery. We plan to make the dataset publicly available in the near future.

The \emph{FaceBase} dataset for craniofacial researchers contains 12,270 3D face scans, but the full cranium is not included.
The only other full head (including full cranium) dataset we are aware of is that of the \emph{Size China} project \cite{Ball2008}.
This tackled head shape for the Chinese population, however, the head fit data remains sparse across other ethnicities and, to our knowledge, statistical models were not developed.

\section{Design and implementation overview}
\label{sec:overview}

\begin{center}
\begin{figure}
\includegraphics[width=1.00\linewidth]{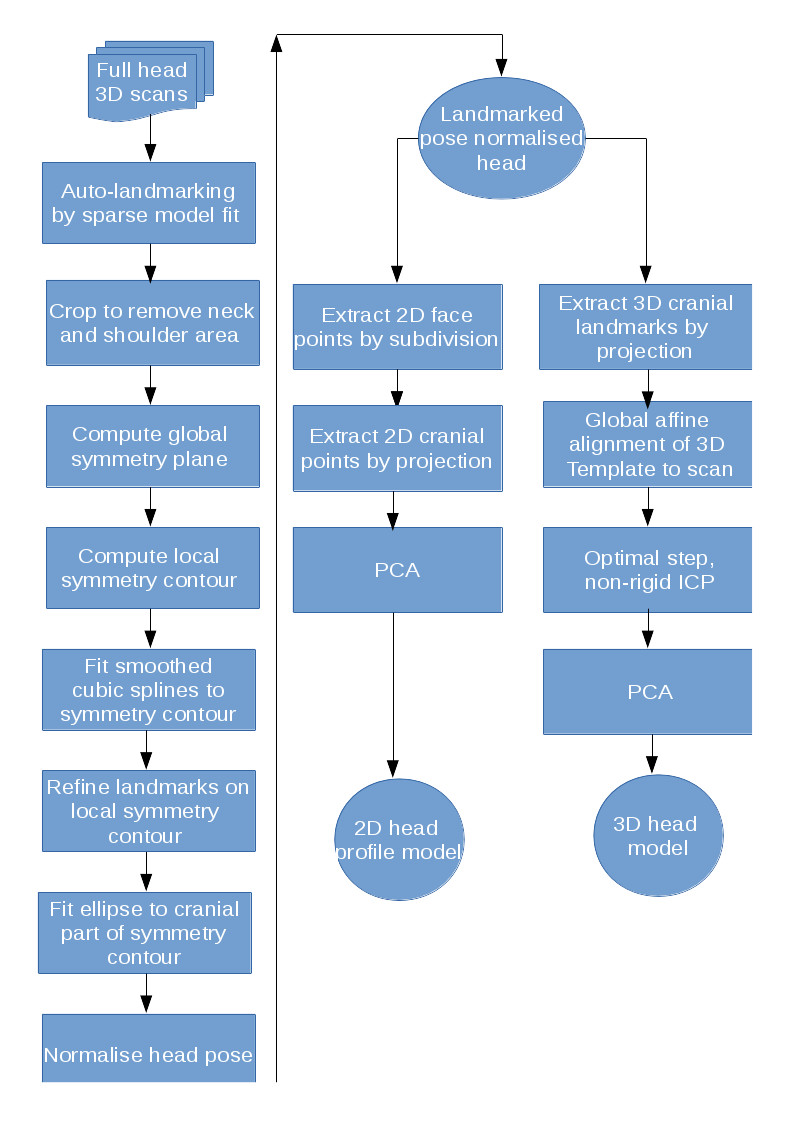} 
\caption{Processes to generate a full head model and profile model}
\label{fig:fhp}
\end{figure}
\end{center}

An overview of our approach is shown in Fig \ref{fig:fhp}. We firstly describe this process in brief in this section, and then in later sections give more detail on each stage, with examples of output results from each stage. Note that the outputs are both a 2D profile model of the face and cranium (close to the bilateral symmetry plane, also known as the sagittal plane) and a full 3D head model.  First we explain the group of processes that generate a landmarked, pose normalised head. These are shown on the left of Fig. \ref{fig:fhp}.

\begin{enumerate}

\item The headspace data presents in arbitrary pose. An automatic landmarking system, trained in a supervised learning scheme, localises 14 facial landmarks and simultaneously generates an \emph{initial} head pose normalisation. An integral part of this automatic landmarking process aligns the 14 landmarks in a least-squares sense to a landmark configuration model, which is positioned in a frontal pose with the mean of its landmark positions at the origin.
\item A scale-adaptive inclined cropping plane is then used to remove the neck and shoulder area of the scan.
\item. The scan is then reflected in the Y-Z plane and a variant of the trimmed Iterative Closest Points (trimmed ICP) algorithm is used to find the global symmetry plane of the head. The trim threshold is gradually reduced to eliminate outliers that do not correspond in a reflection across the symmetry plane.
\item Some regions of the face are detected as asymmetrical, particularly the nose area. Therefore we have developed a local symmetry detection algorithm which, for example, can follow the ridge of the nose, even if the nose points to one side. This is particularly useful for extracting accurate correspondence in head profile modelling.
\item Given a head profile, we fit smoothed cubic splines to give us both a mechanism for extracting points by subdivision and a mechanism for refining landmark positions.
\item Given fitted splines over the local symmetry profile described above, we have developed a local feature refinement method, which optimises various landmark positions that exist on the head profile, such as nasion, pronasale, subnasale and pognion.
\item Since we can extract nasion position on the head profile, we can use this position to segment out the cranium from the full head profile. We have found that, for most heads, it is possible to accurately fit an ellipse to this shape. The centre of this ellipse plays an important role in a new head pose normalisation method.
\item We normalise head pose based on the symmetry plane, ellipse centre position, and nasion position.
\end{enumerate}

Next we follow the set of processes from a landmarked, pose normalised head to the 2D head profile model.

\begin{enumerate}
\item Between each pair of neighbouring landmarks on the symmetry plane profile, we define a fixed number of subdivisions: eg between nasion and pronasale, and we extract points along the spline profile by an interpolation procedure so that the geodesic separation between extracted points is approximately the same. 
\item Points over the cranium are extracted by projecting vectors from the ellipse centre at fixed angular steps (we use one degree).
\item We now have a fixed number of 2D points over the face and cranium 2D profile, so PCA can be applied to generate a morphable 2D profile model.
\end{enumerate}

Finally, we follow the set of processes from a landmarked, pose normalised head to the 3D head profile model.

\begin{enumerate}
\item We extract a set of cranial landmarks by projecting vectors in two orthogonal planes and intersect these vectors with the cranial surface. 
\item These allows a global warping of a 3D head template, which has both face landmarks (the 14 landmarks discussed earlier) and cranial landmarks. 
\item We then apply a variant of Amberg et al's \cite{Amberg07} deformable iterative closest points (ICP) algorithm called {\em optimal step non-rigid ICP}, which was used to build the Basel face Model \cite{bfm09}.
\item Finally we use PCA to generate a 3D head model.
\end{enumerate}

\section{The model building processes in detail}
\label{sec:processes}

In this section we go through each process in more detail and provide example outputs from each stage.

\subsection{Input mesh resolution and mesh reduction}

A sample of 100 adult male head scans (without beards) was selected and the number of vertices checked to three significant figures. All scans were between 106 thousand and 180 thousand vertices, with the mean at 143 thousand and the median at 142 thousand vertices. A single scan was selected with close to the mean number of vertices and the vertex-vertex separation checked over the full mesh. The mean and median vertex separation was 1.19mm. 

We use the MATLAB {\tt reducepatch} function to reduce the vertex and face count  to around 20\% of the original full resolution mesh. The scans now ranged between 20.6k and 37.5k vertices, with the mean at 29.0k vertices and the median at 28.9k vertices. On a sampled scan, the mean vertex separation was 2.86mm and the median 2.72mm. Since the number of vertices reduces by a factor of 5, the mean separation between vertices increases by $\sqrt{5}$. This lower resolution  allows our landmarking processes to run in a reasonable time in a MATLAB implementation. 

\subsection{Landmark-based initial alignment}
\label{sec:landmarking}

\begin{center}
\begin{figure}
\begin{tabular}{cc}
\includegraphics[width=0.4\linewidth]{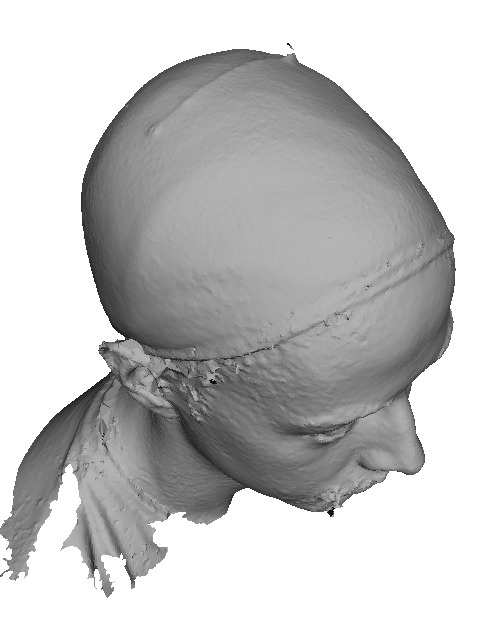} & \includegraphics[width=0.4\linewidth]{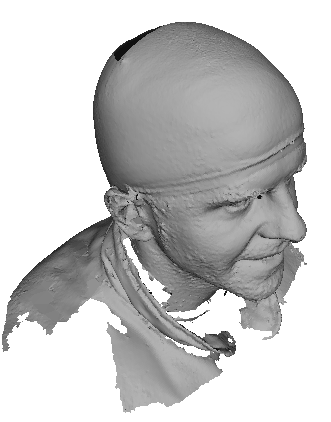}  \\
\end{tabular}
\caption{Two examples of initial pose of full resolution Headspace data.}
\label{fig:headspaceInitial}
\end{figure}
\end{center}

Our system can accept 3D images in arbitrary pose and employs an automatic landmarking scheme to normalise the pose of the 3D head before any further processing. Two examples of typically varying initial pose taken from the headspace datset are shown in Fig. \ref{fig:headspaceInitial}.

\begin{center}
\begin{figure}
\includegraphics[width=1.00\linewidth]{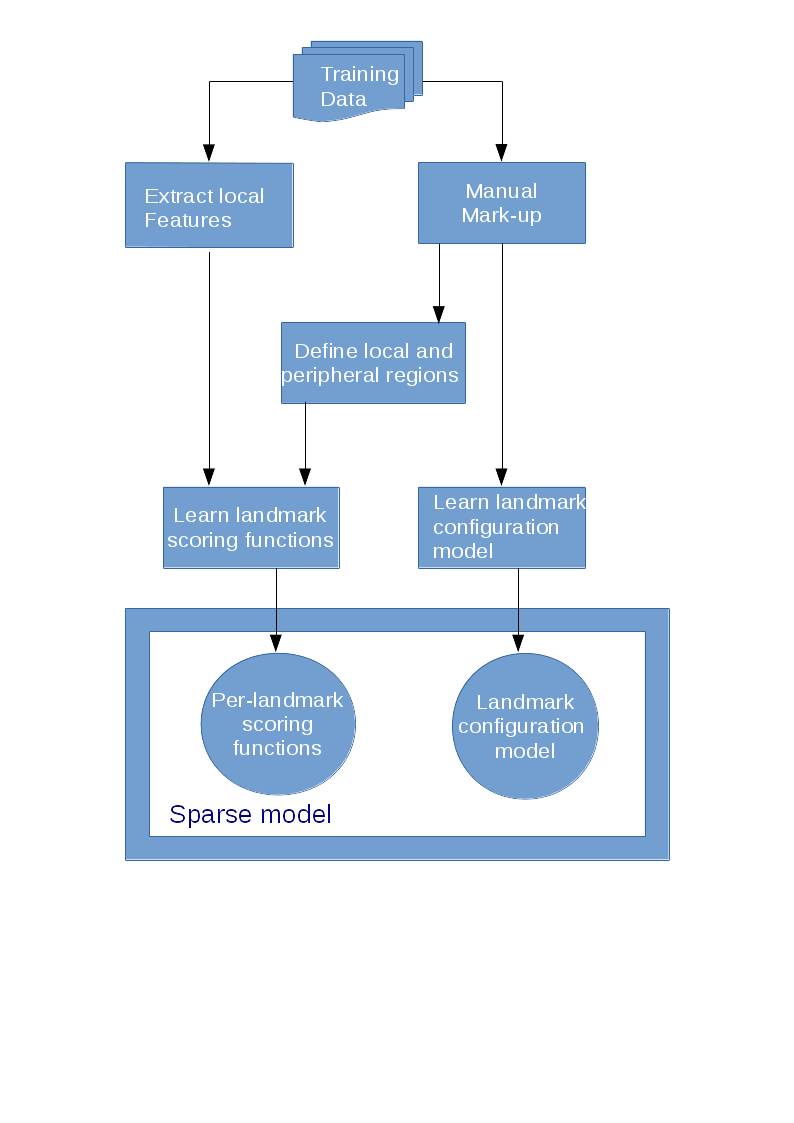} 
\caption{Off-line supervised training of the sparse model for face landmarking}
\label{fig:smTrain}
\end{figure}
\end{center}

\begin{center}
\begin{figure}
\includegraphics[width=1.00\linewidth]{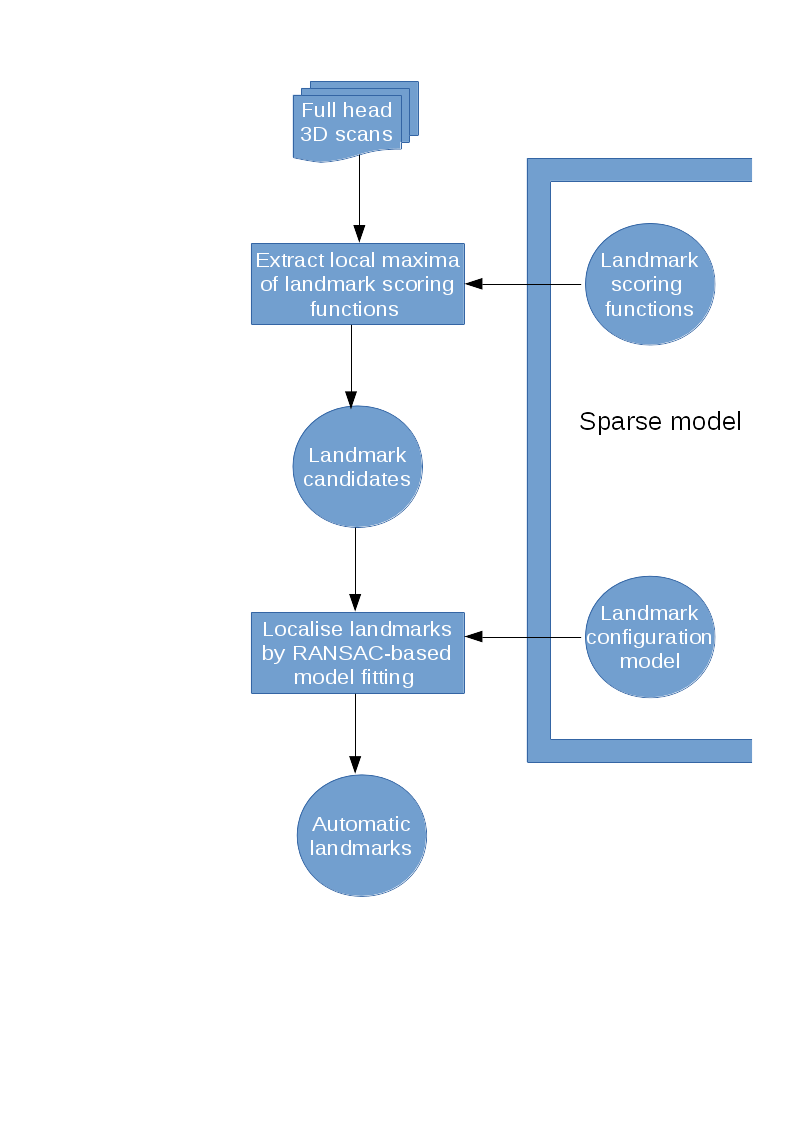} 
\caption{The on-line sparse model fitting process for facial landmarking}
\label{fig:smTest}
\end{figure}
\end{center}

The landmarking system that we have implemented is a modified version of Creusot et al \cite{Creusot2013}, where landmark candidates are generated via per-landmark scoring functions and then a configural model is iteratively fitted using a sample-consensus scheme. There are two parts to this algorithm: an off-line training scheme, shown in Fig. \ref{fig:smTrain} and an online landmarking scheme, shown in \ref{fig:smTest}. Note that we consider a sparse model of the face to consist of a set of landmarks in paticular configuration along with a set of landmark detector functions that score the likelihood of a particular vertex being a particular landmark, based on its local shape. 

\begin{figure}
  \centering
        \includegraphics[width=0.5\linewidth]{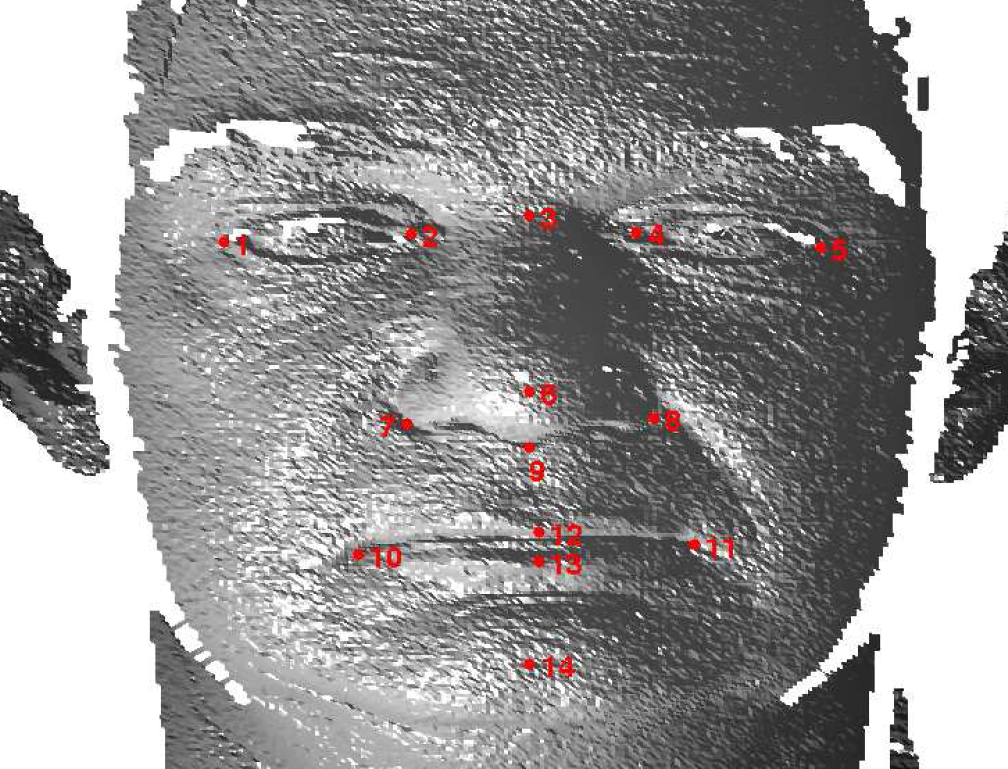}
  \caption{Landmark positions on a sample FRGC scan. Note the surface noise in the raw 3D data. }
  \label{fig:ldmkPos}
\end{figure}

Referring to Fig. \ref{fig:smTrain}, our method is a supervised learning technique - landmarks are marked by hand on a set of training images. We used 279 faces (each different individuals) in the Face Recognition Grand Challenge (FRGC) dataset \cite{Phillips2005}. 
The landmarks used are described in table \ref{tab:ldmks} 
\footnote{Some descriptions taken from www.facebase.org} and can be seen on an example FRGC 3D face scan in Fig. \ref{fig:ldmkPos}.

\begin{center}
\begin{table}
{\small
\vskip 0.5cm
\begin{tabular}{|l|l|l|}
\hline
ID & Name & Description \\ \hline\hline 
1 & Right exocanthion  & Right outer eye corner   \\ \hline
2 & Right endocanthion & Right outer eye corner  \\ \hline
3 & Nasion &  The midline between the orbits, vertically at the \\ 
  &        &  level of the uppermost sulci created by the eyelids.  \\ \hline
4 & Left endocanthion & Left inner eye corner   \\ \hline
5 & Left exocanthion & Left outer eye corner  \\ \hline
6 & Pronasale & The most protrusive point on the nasal tip  \\ 
  &           & in the midline.                             \\ \hline
7 & Right alar curvature point & Junction of the right nasal alare and upper lip   \\ \hline
8 & Left alar curvature point & Junction of the left nasal alare and upper lip   \\ \hline
9 & Subnasale &  The apex of the nasolabial angle in the midline,  \\ 
  &           &  where the inferior border of the nasal septum  \\ 
  &           &  meets the upper lip.  \\ \hline
10 & Right chelion  &  Right mouth corner\\ \hline
11 & Left chelion &  Left mouth corner \\ \hline
12 & Labiale superius &  Centre of upper lip on vermillion line \\  \hline
13 & Labiale inferius  &  Centre of lower lip on vermillion line \\ \hline
14 & Pognion &  Most anterior midpoint of the chin. \\ 
\hline
\end{tabular}
}
\caption{The 14 facial landmarks used in the supervised learning task}
\label{tab:ldmks}
\end{table}
\end{center}

The mean configuration of these landmarks is determined by Generalised Procrustes Analysis (GPA) \cite{DrydenMardia1998} and then put into a standard frontal pose, ensuring that the best fit plane for the symmetry plane landmarks lies in the yz-plane and the subnasale and nasion are at the same depth (z coordinate). 

The local features that we use are spin images \cite{Johnson1997}, and these are computed on FRGC data that has been downsampled by a factor of 16. A spin image can be computed for some mesh vertex ${\mathbf x}$ if we know the normal ${\mathbf{\hat{n}}}$ and the set of neighbouring vertices $\mathbf{x_i}$ within some local spherical neighbourhood. Cylindrical coordinates in the radial direction ($\alpha_i$) and elevation direction ($\beta_i$) are computed as:

\begin{equation}
\beta_i= ( {\mathbf x}-{\mathbf x_i})^T{\mathbf{\hat{n}}},~\alpha_i = \sqrt{(r_i^2 - \beta_i^2)};~ r_i^2 = ( {\mathbf x}-{\mathbf x_i})^T ( {\mathbf x}-{\mathbf x_i})
\end{equation}

\begin{center}
\begin{figure}
\includegraphics[width=0.8\linewidth]{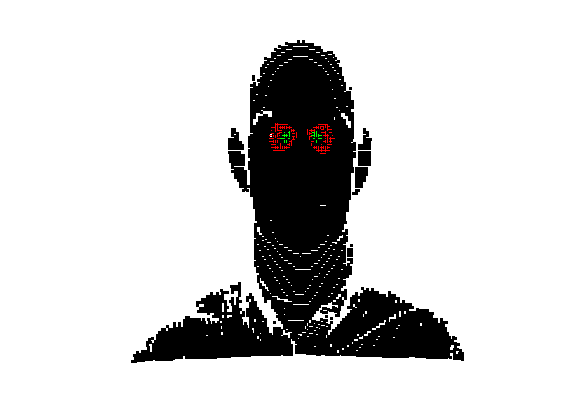} 
\caption{Vertices (on one scan of 279) used to train endocanthion (inner eye corner) landmarking. Landmark vertices (green) are taken to be within 5mm of the manual landmark. Non-landmark vertices (red) are within 10mm and 15mm of the manual landmark.}
\label{fig:trainVerts}
\end{figure}
\end{center}

Values $(\alpha_i, \beta_i)$ for all vertices $\mathbf{x_i}$ are binned into a histogram of dimension $(m,n)$, and we use $m=n=7$.
Radial bins are 3mm in width, giving a local encoding region of radius 21mm. Height bins are 4mm wide and the central bin is centred on zero height to allow more accurate encoding of (near) planar regions. Hence the spin image height range is $\pm$14mm.
(Larger support regions may be more discriminative, but suffer from edge effects in 2.5D datasets, such as FRGC.)
We compress spin images using PCA and discard the dimensions with the smallest variances (smallest eigenvalues), so that
the feature size is reduced from 49 (raw spin images) to 8.
We then learn a scoring function using Linear Discriminant Analysis (LDA) that learns how to distinguish between spin images of vertices near to a particular landmark type and spin images peripheral to that landmark.
The same scoring function is used for symmetrical pairs of landmarks (eg. left and right endocanthions) as spin images are a reflection invariant descriptor (in addition to rotation invariance).
LDA is applied to compressed spin image features and the two classes of vertices in Fig. \ref{fig:trainVerts} are projected onto the direction that best discriminates the two classes. Gaussian density functions are fitted to the 1D distributions of these vertices and the scoring function is computed as the logarithm of the ratio of the landmark density to the peripheral vertex density.

Referring to Fig. \ref{fig:smTest}, we extract features (spin images) across all vertices of the scan, and produce a score map over the full surface for each scoring function. Although there are 14 landmarks, there are only 10 scoring functions as 4 sets of landmrks are paired (left and right endocanthions, left and right exocanthions, left and right mouth corner). The local maxima of each of these scoring functions is stored to generate a set of landmark candidates for each landmark. The pairing of scoring functions means that the same candidates are used for symmetrical landmark pairs. In contrast to using 14 scoring functions \cite{Creusot2013}, this reduces computation and increases the amount of training data for symmetrical landmarks.

Triplets of landmark candidates are used to initialise a scaled alignment (i.e. a similarity transform) between the scan and the configural landmark model (sparse model). Note that the configural model is scaled to match the scan, but the scan is moved into the frame of the sparse model and hence pose normalisation of the scan is inherent in the method. Given that some landmark triplet is used to do an initial alignment, there remains 11 predicted landmark positions that can be used as a consensus measure
to determine if the inital aligning triplet was correct. To form this consensus, we determine the number of landmark predictions that have landmark candidate in their local neighbourhood.
The triplet with the highest consensus, along with its supporting landmarks, is used to form a least squares similarity alignment with the configural model. The nearest vertices on the scans to predicted landmark positions are extracted as the automatic face landmarks on the scan.

As mentioned, we trained the system using 279 distinct subjects (set A, with neutral expression) from the FRGC dataset, which left a further 278 distinct individuals (set B, none appearing in set A) available to test the system. For these, we measured the Euclidean distance between the automatic landmark position and the manual and formed a mean for each landmark; the results are shown in Table. \ref{tab:autoLandmarkingErrors}. The mean over all landmarks was 3.9449mm, which should be interpreted in the context of the low resolution FRGC scans used (downsample factor 16). We analysed the vertex-to-vertex distances across downsampled FRGC test set B to indicate input mesh resolution over which automatic landmarking is performed. The average vertex-to-vertex distance is 3.04mm, and the modal value 2.3mm (when binned over 0.1mm width bins centred at every 0.1mm on the positive reals). Automatic landmarking performance should be interpreted in this context.

\begin{table}
\begin{center}
\begin{tabular}{|l|l|l|} \hline
Ldmk & Name & Mean error (mm) \\ \hline
1 & Outer eye (right) & 4.41094 \\ 
2 & Inner eye (right) & 3.61716 \\ 
3 & Nasion &  2.78653 \\ 
4 & Inner eye (left) & 3.44306 \\ 
5 & Outer eye (left) & 4.38312  \\ 
6 & Nose tip & 3.15003  \\ 
7 & Alare (right) & 2.97366  \\ 
8 & Alare (left) & 2.99983  \\ 
9 & Subnasale & 3.13052  \\ 
10 & Mouth corner (right) & 4.69102 \\ 
11 & Mouth corner (left) & 4.82150 \\ 
12 & Upper centre lip & 3.73908  \\ 
13 & Lower centre lip & 5.04786  \\ 
14 & Chin   & 6.03374  \\ \hline
\end{tabular}
\end{center}
\caption{Mean errors (mm) between auto landmarks and manual landmarks}
\label{tab:autoLandmarkingErrors}
\end{table}

\begin{center}
\begin{figure}
\begin{tabular}{cc}
\includegraphics[width=0.5\linewidth]{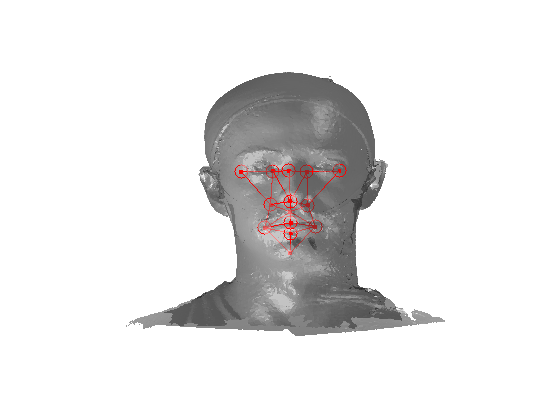} & \includegraphics[width=0.5\linewidth]{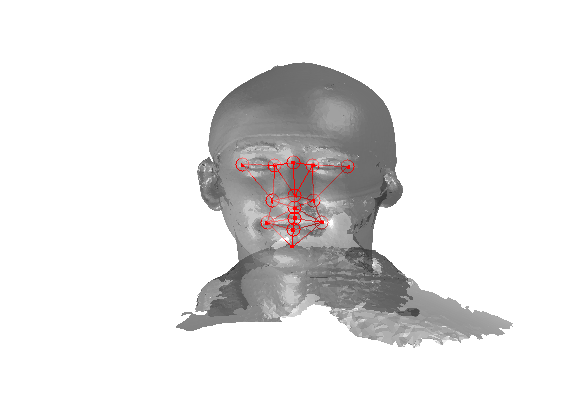} \\
\includegraphics[width=0.5\linewidth]{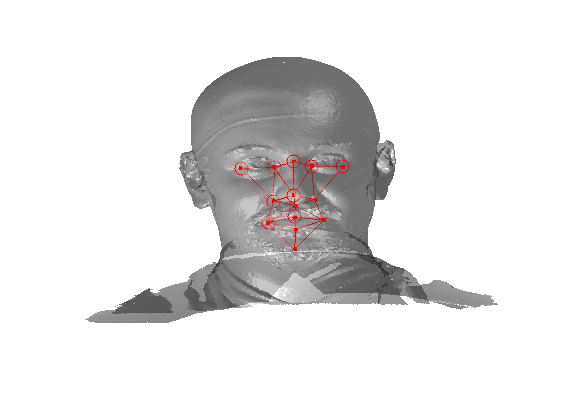} & \includegraphics[width=0.5\linewidth]{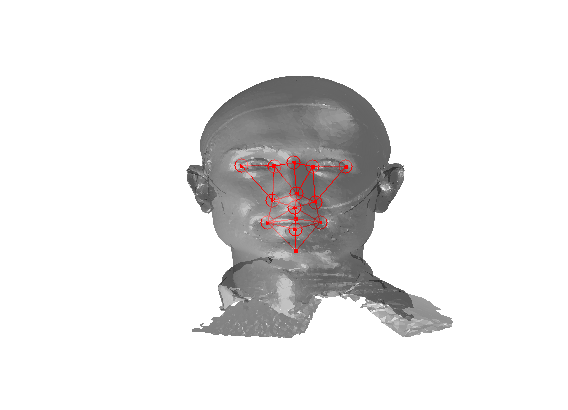} \\
\end{tabular}
\caption{Reduced resolution head space data, pose normalised after the sparse model fit process. Mesh shading is rendered as translucent to highlight the sparse model fit in red. }
\label{fig:smfonhead}
\end{figure}
\end{center}

Fig. \ref{fig:smfonhead} shows examples of the fitted model (trained on FRGC data) on the first four subjects in our \emph{Headspace} training set. Note that the mean of the landmarks in the sparse model is centred on the orgin, and so the scan data is moved towards that and becomes frontal in pose.
We note that other researchers \cite{Wu2014} have used 3D landmark localisation to find the head, but often this only includes landmarks that are relatively easy to localise, namely inner eye corners and nose. Localising more landmarks over a wider facial area has a number of advantage in terms of cropping, pose normalisation and global 3D template warping, all of which are described later.

\subsection{Face and cranium segmentation}

\begin{center}
\begin{figure}
\begin{tabular}{cc}
\includegraphics[width=0.5\linewidth]{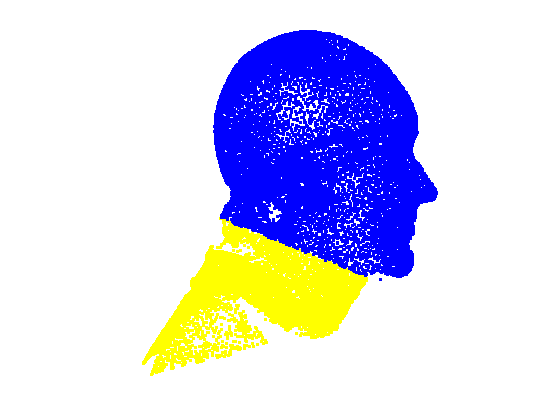} & \includegraphics[width=0.5\linewidth]{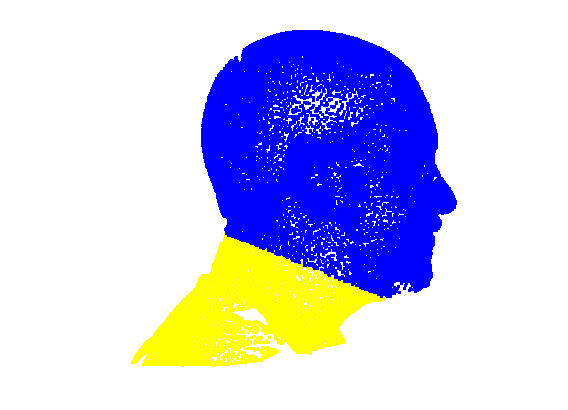}  \\
\end{tabular}
\caption{A head cropping plane, positioned relative to the sparse model fit is used to segment the face and head region from the neck and shoulder region. Blue vertices are retained and yellow discarded.}
\label{fig:croppingPlane}
\end{figure}
\end{center}

Fitting a sparse model to the head scan data normalises its pose to frontal. Since we use landmarks that localise the majority of the
facial area (rather than just inner eyes and nose), we can reliably position a scale-adaptive mesh cropping plane. We choose a distance below the chin landmark, that is some proportion of the face length, defined  by the distance from the nasion landmark to the chin landmark.
This gives a cropping plane that scale-adapts to the size of the head and it is aligned such that it is at some fixed angle (eg. 40 degrees, chosen experimentally) in the Y-Z plane. This is then used to remove the neck and shoulder area of the scan, as shown in Fig. \ref{fig:croppingPlane}. 

\subsection{Global symmetry plane extraction}

The scan is reflected in the Y-Z plane and a variant of the trimmed Iterative Closest Points (trimmed ICP) algorithm is used to find the global symmetry plane of the head. This \emph{reflect and register} approach was first proposed by Benz et al \cite{Benz2002} to find the symmetry plane of 3D face data. Here, pose normalisation resulting from the sparse model fitting stage ensures that ICP always converges correctly. Additionally, the use of a trim threshold that is gradually reduced removes outliers that do not correspond to a reflection across the symmetry plane.
 
\begin{center}
\begin{figure}
\begin{tabular}{cc}
\includegraphics[width=0.4\linewidth]{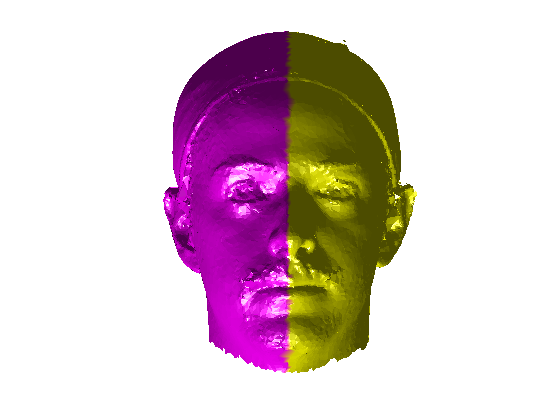} & \includegraphics[width=0.4\linewidth]{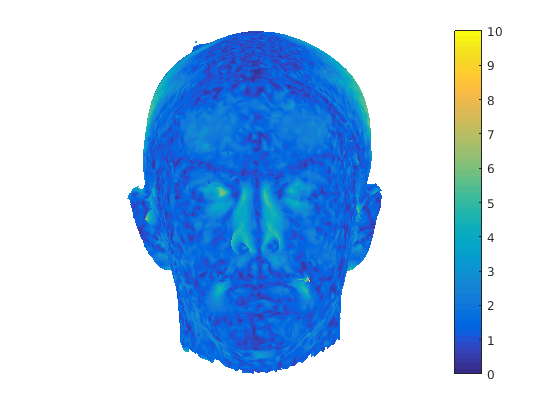} \\
\includegraphics[width=0.4\linewidth]{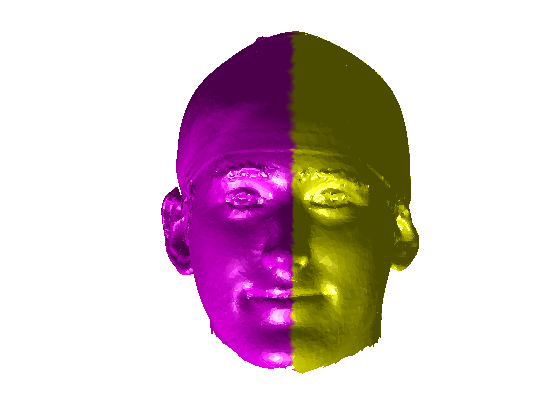} & \includegraphics[width=0.4\linewidth]{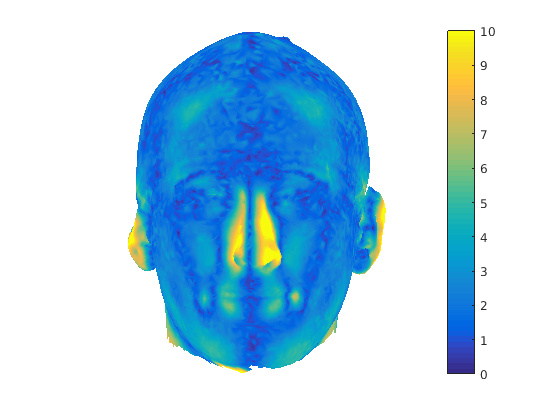} \\
\includegraphics[width=0.4\linewidth]{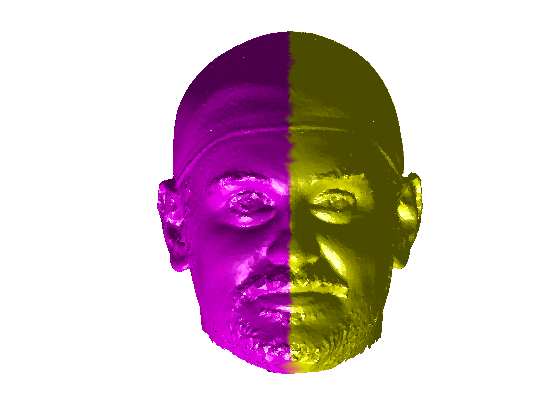} & \includegraphics[width=0.4\linewidth]{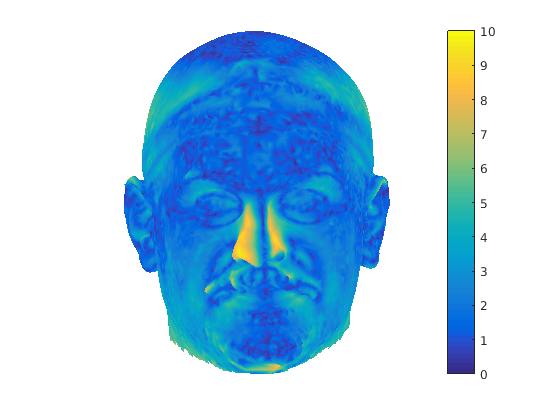} \\
\includegraphics[width=0.4\linewidth]{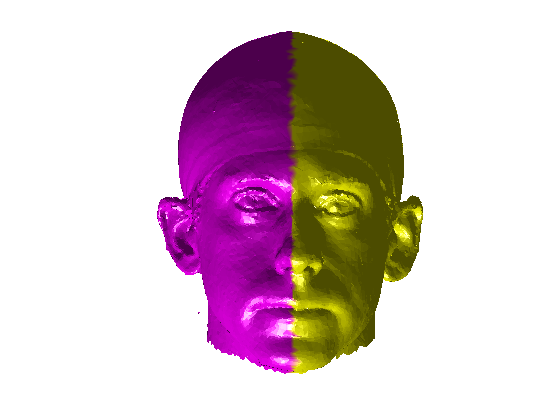} & \includegraphics[width=0.4\linewidth]{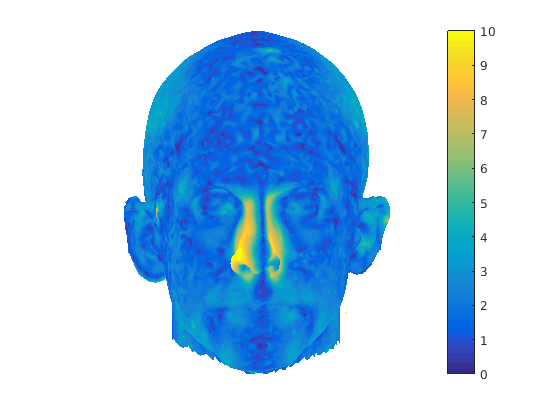} 
\end{tabular}
\caption{Left images: global symmetry plane of the head is used to colour the left and right of each image. Right images: colourmap of the Euclidean distance between nearest neighbours on the scan and its reflection. Notice that regions around the nose are often rejected by thresholding in the trimmed ICP process.}
\label{fig:gsp}
\end{figure}
\end{center}

The region that the head scan and its reflection lie in is pre-partitioned into cuboid bins to reduce the computational cost of nearest neighbour search. Fig. \ref{fig:gsp} show the results of global symmetry plane alignment for 8 adult male samples from the headspace dataset. From observing 100 of these scans we have noticed that the nose, a cartilaginous structure, is often not aligned with the global symmetry plane and regions around the nose area are often rejected as outliers in the trimmed ICP process. As a result, it is not possible to build a good profile model of the face using global symmetry alone, as the ridge of the nose is often not sufficiently well aligned with the global symmetry plane. This motivated us to find the facial contour using local symmetry considerations.

\subsection{Facial contour extraction using local symmetry}

\begin{center}
\begin{figure}
\begin{tabular}{cc}
\includegraphics[width=0.4\linewidth]{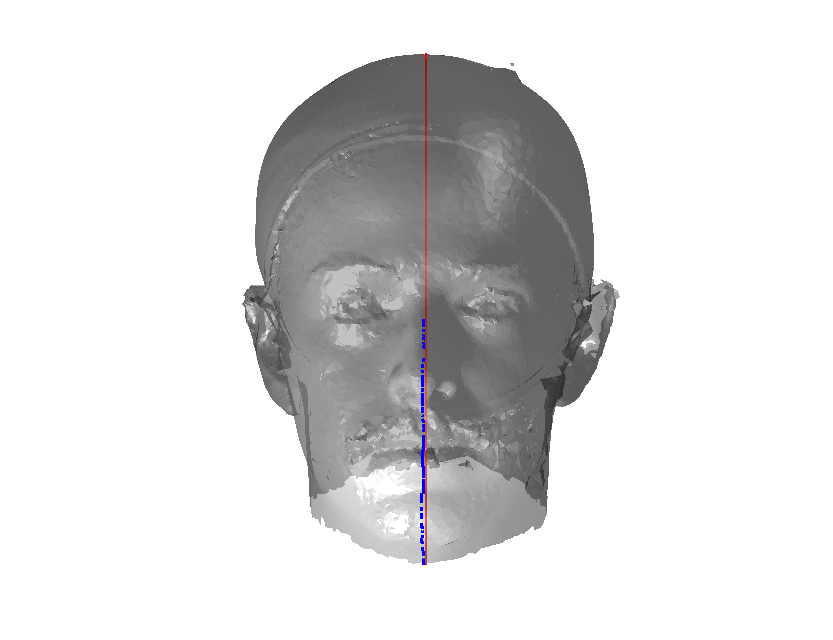} & \includegraphics[width=0.4\linewidth]{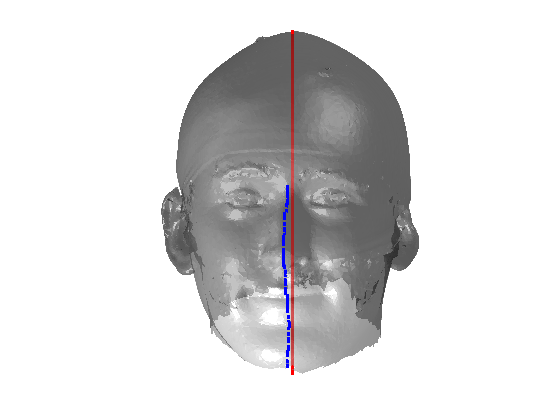} \\
\includegraphics[width=0.4\linewidth]{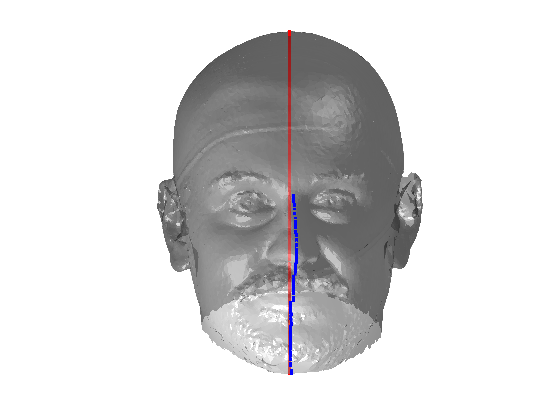} & \includegraphics[width=0.4\linewidth]{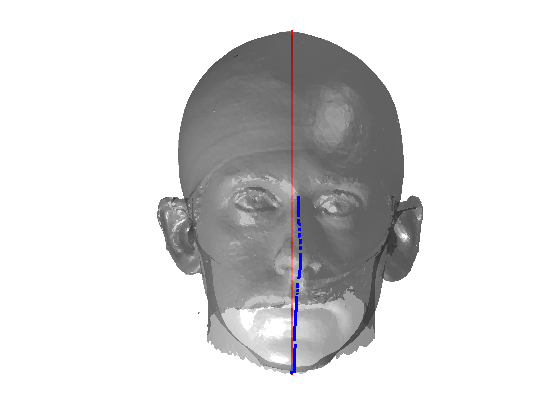} 
\end{tabular}
\caption{Facial local symmetry contours (dark blue) deviating from the global symmetry plane (red).}
\label{fig:lsc}
\end{figure}
\end{center}

In order to find the facial contour we implemented a piecewise ICP process across the glocal symmetry plane.
Each head is first rotated such that its global symmetry plane is coincident with the Y-Z plane. 
The facial region, as defined by the sparse model fit is then divided into a set of horizontal strips (we use 20mm steps between the nasion and pognion) in the $y$ dimension, with the back of the head being cropped out. We apply our trimmed ICP algorithm to each separate strip, so that a local symmetry plane is found for each strip and the facial contour is found by intersecting this sequence of local symmetry planes with its corresponding facial strip. For each strip (index $i$), we use the strip below $(i-1)$ and the strip above $(i+1)$ in the trimmed ICP process in order to reduce the sensitivity of the local symmetry planes to noise. This process is only applied to the face and the cranial region is dealt with as a whole. This is because, in strips, it does not provide sufficient constraints for ICP to lock onto and the two surfaces can slide over each other freely. Fig. \ref{fig:lsc} shows facial local symmetry contours deviating from the global symmetry plane.
Points on the head profile are then extracted by detecting mesh arcs that cross a symmetry plane and linearly interpolating to extract 3D vertices that lie on that symmetry plane.

\subsection{Cubic spline fitting}

\begin{center}
\begin{figure}
\begin{tabular}{cc}
\includegraphics[width=0.4\linewidth]{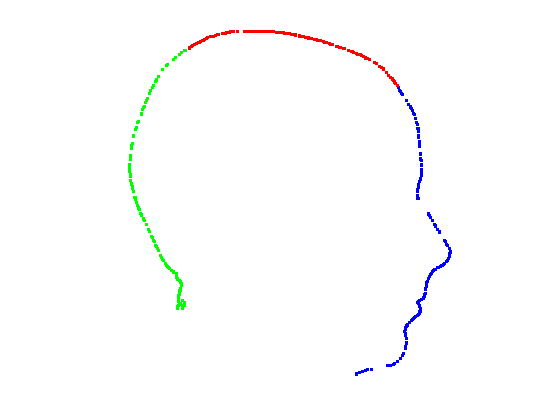} & \includegraphics[width=0.4\linewidth]{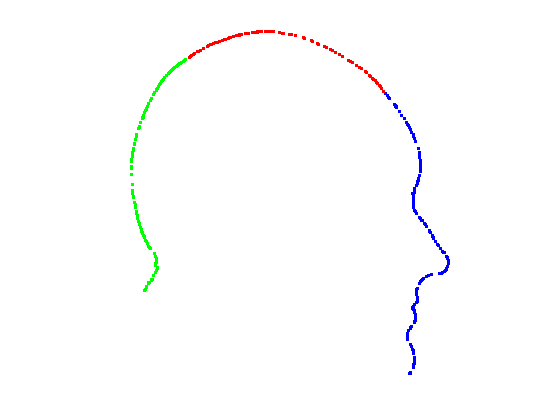} \\
\includegraphics[width=0.4\linewidth]{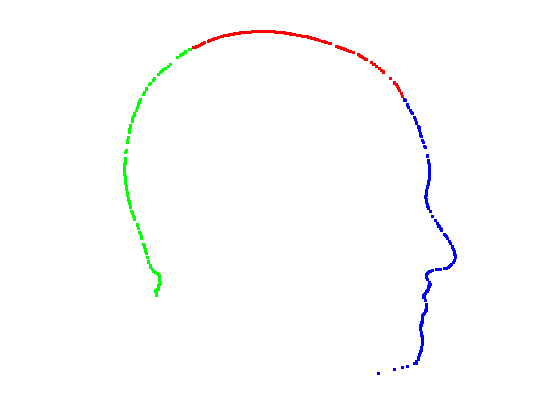} & \includegraphics[width=0.4\linewidth]{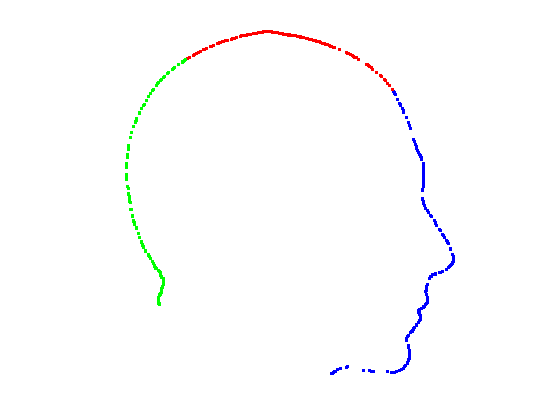} 
\end{tabular}
\caption{Points intersecting symmetry planes are extracted and segmented into three groups, as shown by the red, green and blue colours.}
\label{fig:seg}
\end{figure}
\end{center}

Given a set of points that lie on a symmetry plane, we divide the points into three groups by segmenting them at angles of 45 degrees and 135 degrees relative to the origin, as shown in Fig. \ref{fig:seg}. This gives us three profiles that we can model with spline functions.

\begin{center}
\begin{figure}
\begin{tabular}{cc}
\includegraphics[width=0.4\linewidth]{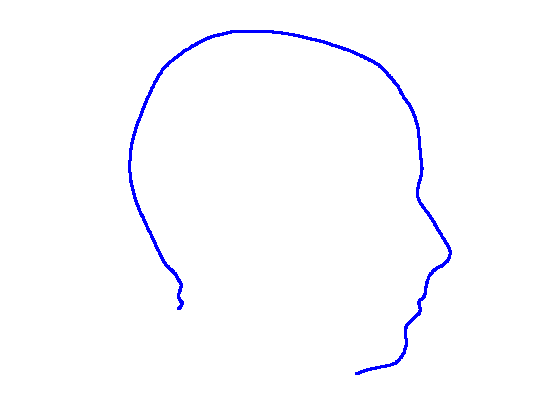} & \includegraphics[width=0.4\linewidth]{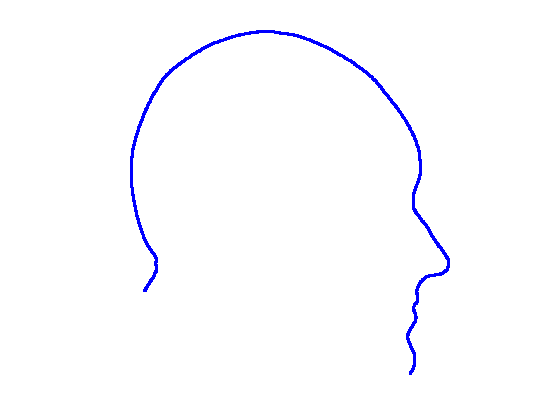} \\
\includegraphics[width=0.4\linewidth]{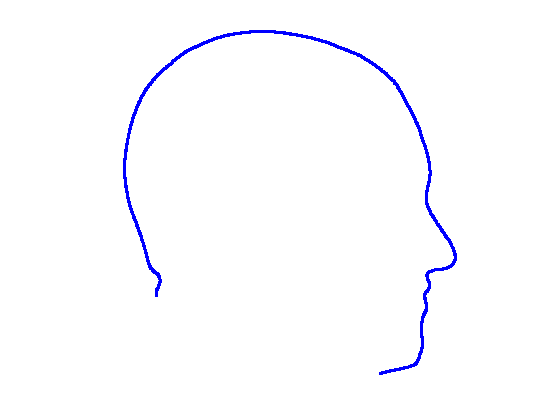} & \includegraphics[width=0.4\linewidth]{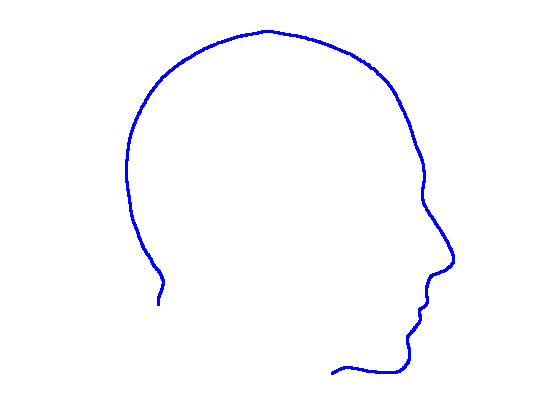} 
\end{tabular}
\caption{Profiles generated by fitting cubic splines to three segments.}
\label{fig:splines}
\end{figure}
\end{center}

For each of the three segments, we determine the range of gradients on that segment by simple differencing between consecutive points. Each segment is rotated into a 2D frame such that maximum absolute values of positive and negative gradients are equal. Cubic spline contours are fitted to the three profiles and the head profile contour is reconstructed using the three individual cubic splines. Fig. \ref{fig:splines} show examples of these.

\subsection{Profile feature refinement}

A machine learning method of finding approximate facial landmark localisations was described in section \ref{sec:landmarking}.
The $(y,z)$ positions of these landmarks are used to indicate the approximate locations of landmarks on the extracted profile, by closest point projection. These can be considered as initial approximate landmark positions, which then can be refined by a local optimisation.

The optimisation method that we designed is based on what we term a \emph{disc operator}. In this method, we centre a disc (at the largest scale that we are interested in) on some point on the head profile and fit, by least squares, a quartic polynomial to the profile points within that disc. A quartic was chosen to give the flexibility to fit to the `double peaked' area over the lips. Thus we find quartic parameters $p^T$ to fit a set of profile points $[\mathbf{x_p},\mathbf{y_p}]$ such that, with $n=4$:

\begin{equation}
\hat{y_p} = {\mathbf p}^T {\mathbf x_p}, ~~~ {\mathbf p} = [p_0 ... p_n]^T, {\mathbf x_p} = [x_p^0 ... x_p^n]^T
\end{equation}

To implement our disc operator, we take a dense, regularly spaced set of $n$ point samples within that disc, $[{\mathbf x_d}, {\mathbf y_d}]$ and compute the operator value as

\begin{equation}
\alpha = \frac{1}{n}\displaystyle\sum_{i=1}^{n} sign( y_d - {\mathbf p}^T {\mathbf x_d}  )
\label{eqn:discoperator}
\end{equation}

We find that $( -1 \le \alpha \le 1 )$, with values close to zero indicating locally flat regions, positive values indicating convexities, such as the pronasale (nose tip), and negative values indicating concavities, such as the nasion.
Effectively this is a discrete approximation to finding the area within the facial profile that intersects with a disc of some predefined scale. The discrete sampling gives a high frequency quantisation noise on the signal and so we filter this with a 10th order low pass Butterworth filter. 

\begin{center}
\begin{figure}
\begin{tabular}{cc}
\includegraphics[width=0.5\linewidth]{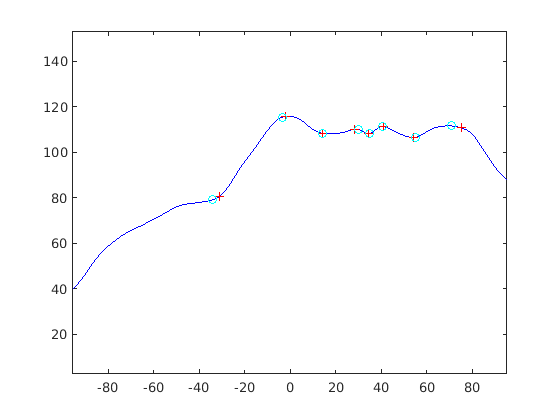} & \includegraphics[width=0.5\linewidth]{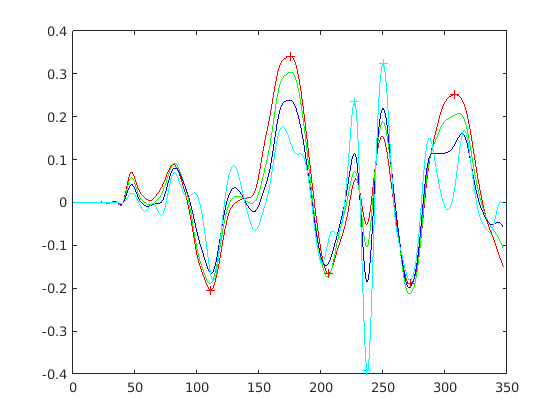} \\
\end{tabular}
\caption{Left: Automatic (red cross) and manual (cyan circle) landmarks on a facial profile. Right: the disc operator output at multiple scales, with landmark positions at the signal's extrema. Red = 20mm radius, green = 15mm, blue = 10mm, cyan = 5mm radius. Landmarks are located at the strongest scale, which is different for different sized features.  }
\label{fig:cpLdmks}
\end{figure}
\end{center}

The operator is very easily adapted to multiple scales, with smaller scales being straightforward subsamples of the disc points $[{\mathbf x_d}, {\mathbf y_d}]$ at larger scales. In this case, we reapply Eqn. \ref{eqn:discoperator} to the subsample and, as with the larger scale, apply a low pass filter. A useful property of the operator, in contrast to a standard curvature operator, is that its scale is easily tuned to the size of the features that one is interested in: we use a 20mm disc radius for most profile landmarks (nasion, pronasale, subnasale, chin concavity and pognion), except those around the lips (upper lip, lower lips, centre of lips), where we use the smallest scale, 5mm. Fig. \ref{fig:cpLdmks} illustrates the disc operator's output at four scales to illustrate its behaviour.
For 5 of 8 facial profile landmarks, the strongest output is (usually) the largest scale. For the remaining three, the strongest output is the smallest scale.

The landmarking algorithm employed finds the nearest strong local extrema of the appropriate sign and at the appropriate scale.
Our initial thought was to use a search from the largest to the smallest scale for all landmarks, We implemented this, but it often failed around the lip region, perhaps because the lower lip and upper lip localisations in the initial sparse model fit are very close to each other. Instead, we refine the pognion (chin), subnasale, pronasale (nose tip) and nasion at the largest operator scale. We then
consider upper, centre and lower lips simultaneously by looking for a strong M-shaped disc operator signature at the smallest scale (5mm, cyan, in Fig. \ref{fig:cpLdmks}), between the subnasale and pognion. Finally, we find the chin cleft location as the strongest minimum between the lower lip and pognion.

To evaluate our feature optimisation, we manually marked 100 training scans with eight landmarks. These include six landmarks related to the inital sparse model fit, and a further two, one of which is the concavity between upper and lower lips and the other is a concavity defining the cleft between the lower lip and chin. The results of the mean error, relative to the manual markup and the standard deviation, representing the variation relative to the manual markup, is given in Table \ref{tab:discopResults}.

\begin{table}
\begin{center}
\begin{tabular} {|l|l|l|l|l|} \hline 
Ldmk \# & Ldmk name & Mean (mm) & SD (mm) & Max (mm) \\ \hline \hline 
1 & pognion (chin) & -3.796 & 3.138 &  1.805 \\ \hline 
2 (*) & chin cleft & 0.237 & 1.254 &  6.952 \\ \hline 
3 & lower lip & -0.576 & 0.827 &  1.855 \\ \hline 
4 (*) & centre lip & -0.317 & 0.534 &  0.760 \\ \hline 
5 & upper lip & -0.001 & 0.853 &  1.499 \\ \hline 
6 & subnasale & 0.992 & 1.195 &  4.306 \\ \hline 
7 & pronasale & -2.011 & 0.990 &  0.000 \\ \hline 
8 & nasion & -1.876 & 1.031 &  0.679 \\ \hline 
\end{tabular}  
\caption{Mean, standard deviation and maimum  of disc operator landmark localisation errors relative to manual for 100 3D face scans. Those marked with a * were not present in the original sparse model fit.}
\label{tab:discopResults}
\end{center}
\end{table}

A non zero mean value in the error indicates a bias of the automatic localisation relative to manual localisation. 
In terms of model building, this may have a significantly lower effect on the performance of the model
than the standard deviation figure in Table \ref{tab:discopResults}. We note, for example, that the operator consistently marks the nasion lower than a human operator, but the standard deviation is small, compared to that of the chin. We will evaluate the compactness of models build from manual landmarks and automatic landmarks later in Sect. \ref{sec:compactness}. 

\subsection{Ellipse fitting}

Ellipse fitting was motivated by the fact that large sections of the cranium appeared to be elliptical in form, 
thus suggesting a natural centre and frame origin with which to model cranial shape. An ellipse belongs to the set of curves known as conics, which can be defined by the equation:

\begin{equation}
{\mathbf a} . {\mathbf x} = 0, ~ {\mathbf a} = [a~b~c~d~e~f]^T, {\mathbf x}= [x^2~xy~y^2~x~y~1]^T
\end{equation}

\noindent
and if $\mathbf a$ is constrained such that the discriminant $b^2-4ac$ is negative, then the conic is an ellipse.
We use the nasion landmark position to segment out the cranium region from the face and use a robust iterative ellipse fitting procedure that rejects outliers. The core of this algorithm is the \emph{Direct Least Squares Fitting of Ellipses} approach of Fitzgibbon et al \cite{Fitzgibbon99}, which presents a solution for $\mathbf a$ as the solution of an eigensystem that enforces the required constraint on the discriminant.

However, the algorithm in its raw form is insufficient as we need to reject outliers, as is often caused by the latex cap peak worn by imaged subjects. Thus we use an iterative call to the procedure in \cite{Fitzgibbon99} that rejects outliers outside some threshold of the fitted ellipse contour, iterating until the inlier set becomes stable or a maximum number of iterations is reached.

\begin{center}
\begin{figure}
\begin{tabular}{cc}
\includegraphics[width=0.4\linewidth]{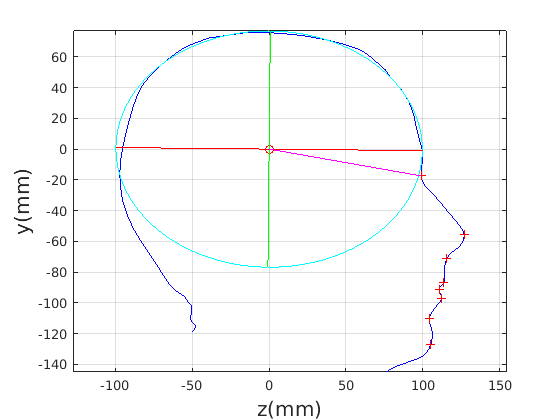} & \includegraphics[width=0.4\linewidth]{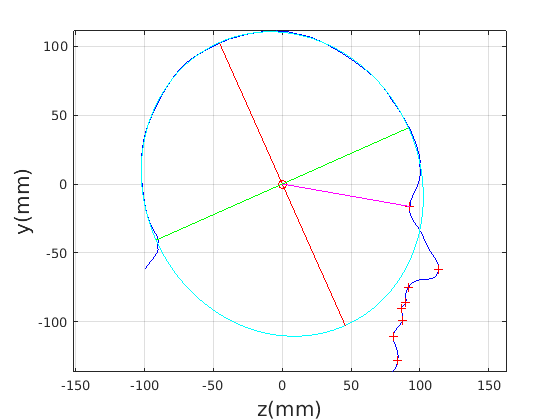} \\
\includegraphics[width=0.4\linewidth]{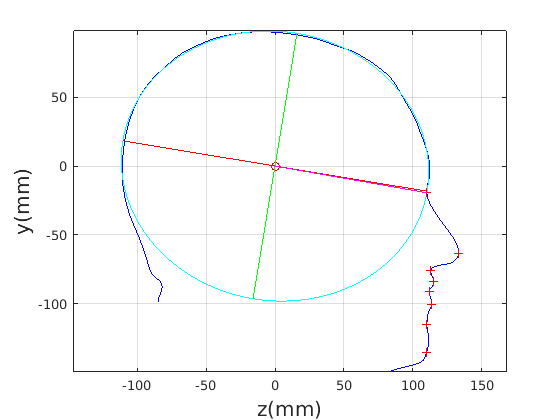} & \includegraphics[width=0.4\linewidth]{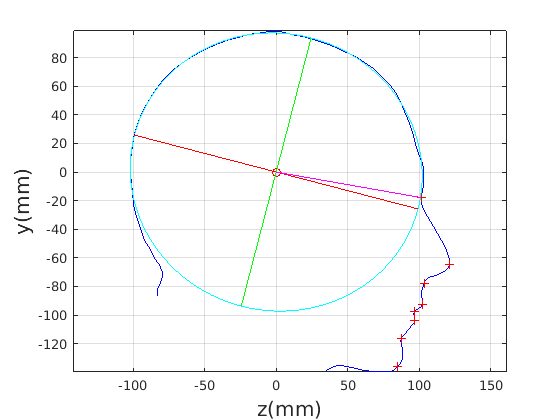} 
\end{tabular}
\caption{Head tilt pose normalisation based on ellipse centre and nasion position. The extracted head profile is shown in blue, red crosses show facial landmarks and the ellipse fitted to the cranial profile is shown in cyan. Its major axis is red and its minor axis green. The head profile is centred (translated) onto the ellipse centre and then rotated such that the naison is at -10 degrees from the horizontal. This is indicated by the magenta line on the figure.}
\label{fig:ellipseFit4}
\end{figure}
\end{center}

Figure \ref{fig:ellipseFit4} shows examples of the robust ellipse fit for four head profiles. The centre of the ellipse is used in a pose normalisation procedure where the ellipse centre is used as the origin of the profile and the angle from the ellipse centre to the nasion is fixed at -10 degrees. We call this Ellipse Centre - Nasion (ECN) pose normalisation and later compare this to GPA. The major and minor axes of the extracted ellipses are plotted as red and green lines respectively in Fig. \ref{fig:ellipseFit4}.

\begin{center}
\begin{figure}
\includegraphics[width=1.0\linewidth]{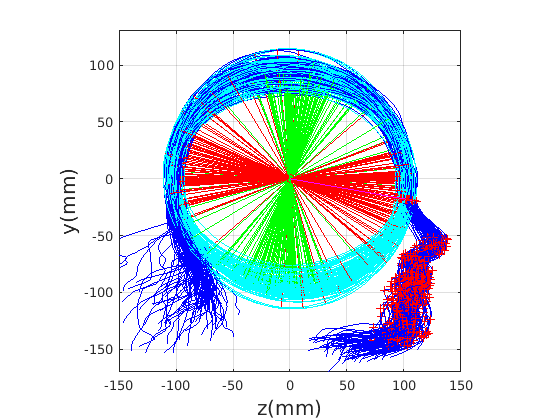}   
\caption{All 100 training profiles overlaid after ECN normalisation.}
\label{fig:ellipseFitAll}
\end{figure}
\end{center}

\begin{figure}
\centerline{\includegraphics[width=0.8\linewidth]{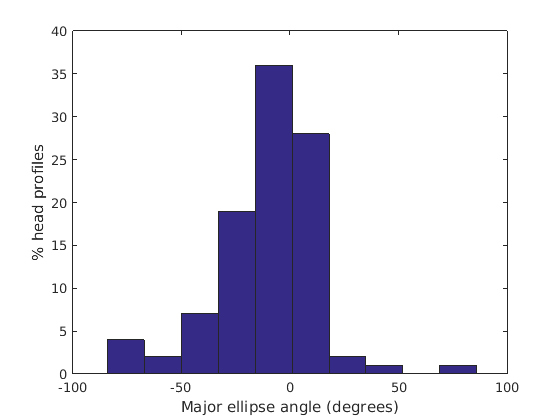} }  
\caption{Major axis ellipse angles with respect to an ECN baseline of -10 degrees: median angle is -6.4 degrees (2sf).}
\label{fig:ellipseAngles}
\end{figure}

Figure \ref{fig:ellipseFitAll} shows all 100 profiles overlaid with the same alignment scheme. The median value of major ellipse axis and the ellipse centre-nasion angle differ by 3.6 degrees, so that when the nasion angle is fixed at -10 degrees, the median ellipse angle is -6.4 degrees (cf. -7.4 degrees with manual landmarking of the nasion). We noted regularity in the orientation of the fitted ellipse as is indicated by the clustering of the major (red) and minor (green) axes in Fig.
\ref{fig:ellipseFitAll} and the histogram of ellipse orientations in Fig. \ref{fig:ellipseAngles}. 
For most people, the major axis of the ellipse is closely aligned with the y-axis (upright), and titled slightly forwards.  
A minority of heads (9\%) in the training sample have their major ellipse axes closer to the vertical (these relatively tall and short heads are known as brachycephalic.)
Ellipse axis clustering (relative to the fixed ellipse centre-nasion line) does not appear to be sharply defined. 
This is because many crania are close to circular in cross-section, making the orientation of these angles sensitive to small changes in shape from one person to the next. Note also the variation at the back of the head due to a variety of hair styles, some of which protrude from under the cap. We limit the region over which we model the cranial shape in order to crop this unwanted data out.

\subsection{Extracting profile model points: curve subdivision and vector projection.}

To extract profile points that will be part of our morphable profile model, we have an interpolation procedure that ensures that there is a fixed number of evenly-spaced points between any pair of face profile landmarks. 

To achieve this we consider the cubic spline fit to be piecewise linear (a good approximation for small enough steps along the spline) 
and we compute the cumulative along-contour distance between some pair of landmarks as we move from one spline contour point to the next. Given a required $n$ points between some landmark pair, the approximate geodesic distance between them is divided into $(n+1)$ equal length segments to generate required geodesic distances for each new profile point. For each of these we determine the pair of cubic spline points that the new point lies between and we determine its position by linear interpolation.


It is not possible to landmark the cranial region and extract profile model points in the same way as the facial area.
This area is smooth and approximately elliptical (usually close to circular) in structure. Instead we choose to project vectors from the ellipse centre and intersect the cubic spline curves , starting at the nasion, and incrementing the angle anticlockwise in small steps (we use one degree) over a fixed angle. (We use 210 degree to stop short of the bottom of the latex cap, where hair often protrudes.)

\subsection{Sagittal plane head profile models}

Our head profile models are built by pose normalisation of $n=100$ adult male sagittal profiles in the training set followed by the application of Principal Component Analysis (PCA). Each profile is represented by $m$ 2D points $(y_i,z_i)$ and is reshaped to a $2m$ row vector. Each of these vectors is then stacked in a $n \times 2m$ data matrix, and each column is made zero mean. Singular Value Decomposition is applied from which eigenvectors are given directly and eigenvalues can be computed from singular values. This yields a linear model as:

\begin{equation}
\mathbf{x_i} = \mathbf{\bar{x}} + \mathbf{P b_i} = \mathbf{\bar{x}} +  \sum_{i=1}^{k} \mathbf{p^k} b^k_i
\end{equation}

\noindent
where $\bar{x}$ is the mean head profile shape vector and $\mathbf P$ is a matrix whose columns $\mathbf p^k$  are the eigenvectors of the covariance matrix (after pose alignment), describing orthogonal modes of head profile variation. The vector $\mathbf b$ holds the shape parameters $\{b^k\}$, that weight the shape variation modes which, when added to the mean shape, model a shape instance $\mathbf x_i$.
In our case $2m > n$ and so the maximum number of modes of shape variation is $k_{max} = n-1$ although, in practice,
model compactness is such that only around 5\% of the maximum number of shapes modes are required to accurately model shape variations.

\section{Model evaluation}
\label{sec:evaluation}

\subsection{Modes of form and shape variation}

We first demonstrate the dominant modes of shape variation over our training set. This is done for:

\begin{enumerate}
\item The craniofacial profile : no scale normalisation (a form model).
\item The craniofacial profile : with scale normalisation (a shape model).
\item The cranial profile : no scale normalisation (a form model).
\item The cranial profile : with scale normalisation (a shape model).
\end{enumerate}

\begin{center}
\begin{figure}
\begin{tabular}{cc}
\includegraphics[width=0.4\linewidth]{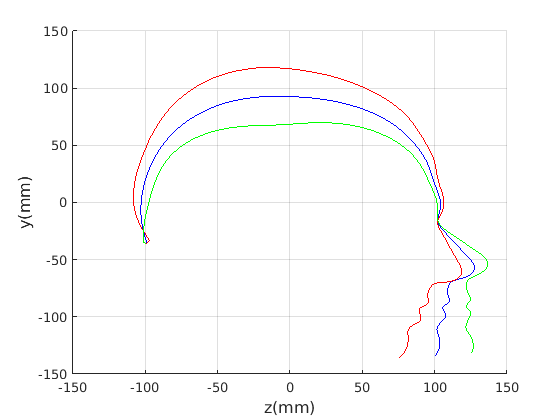} & \includegraphics[width=0.4\linewidth]{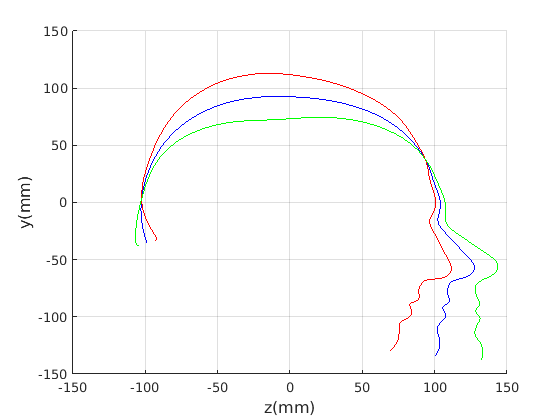} \\
\includegraphics[width=0.4\linewidth]{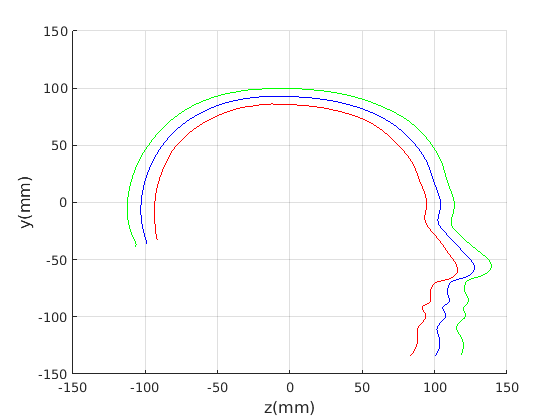} & \includegraphics[width=0.4\linewidth]{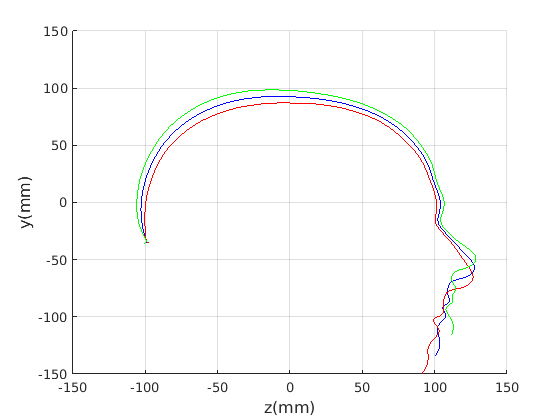} \\
\includegraphics[width=0.4\linewidth]{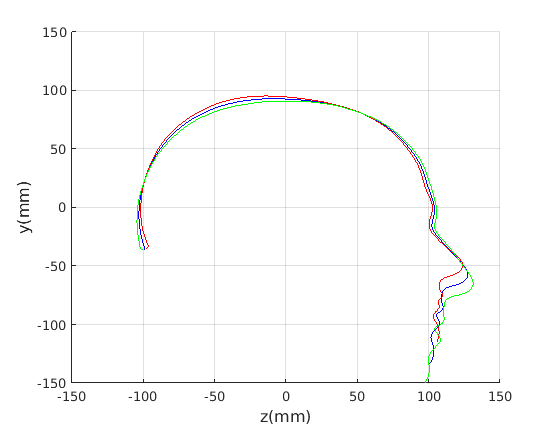} & \includegraphics[width=0.4\linewidth]{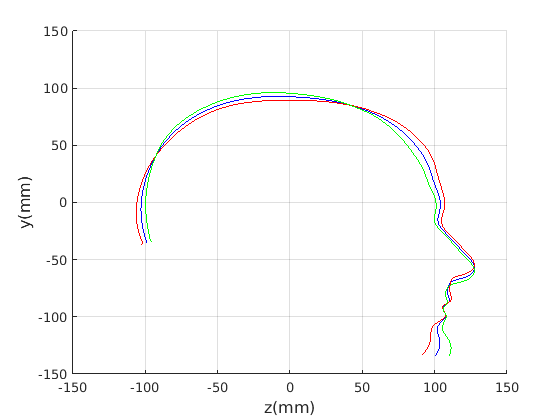} \\
\includegraphics[width=0.4\linewidth]{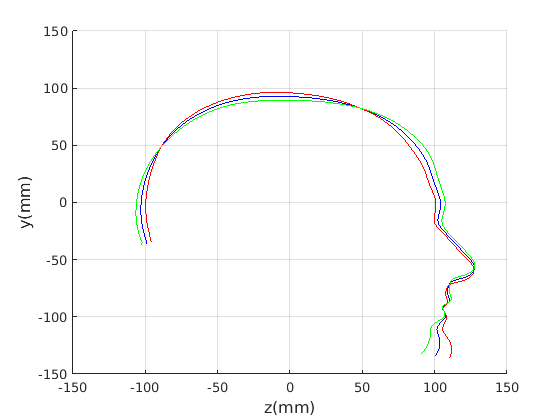} & \includegraphics[width=0.4\linewidth]{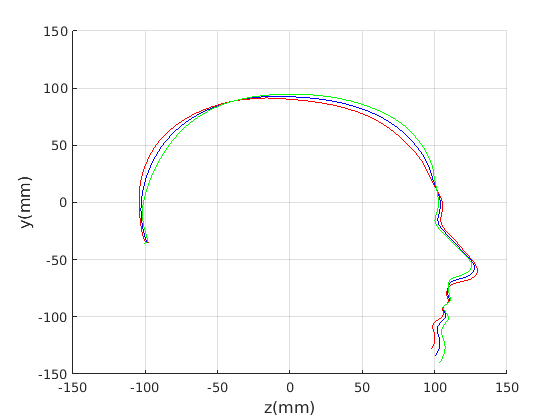} 
\end{tabular}
\caption{Left column: The dominant four modes (Top: mode 1; bottom: mode 4) of head shape variation using automatic profile landmark refinement and ECN normalisation. Mean is blue, mean+3SD is red and mean-3SD is green. Right column : scale normalised variant.}
\label{fig:headProfECN}
\end{figure}
\end{center}

\begin{center}
\begin{figure}
\begin{tabular}{cc}
\includegraphics[width=0.4\linewidth]{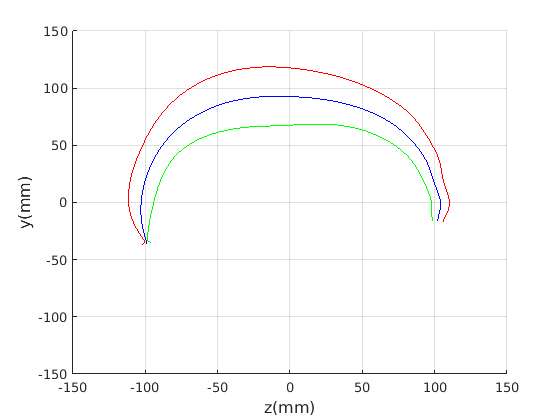} & \includegraphics[width=0.4\linewidth]{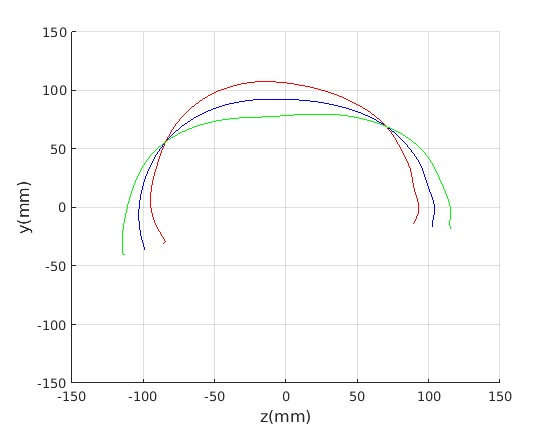} \\
\includegraphics[width=0.4\linewidth]{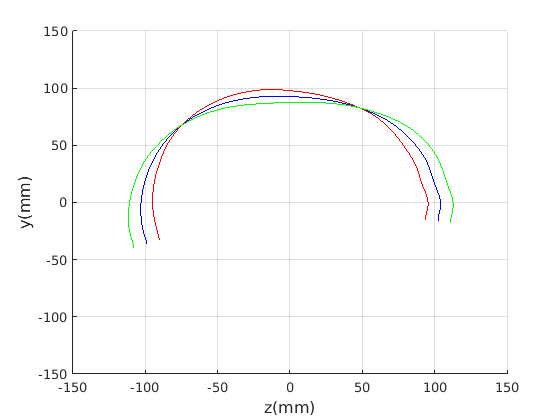} & \includegraphics[width=0.4\linewidth]{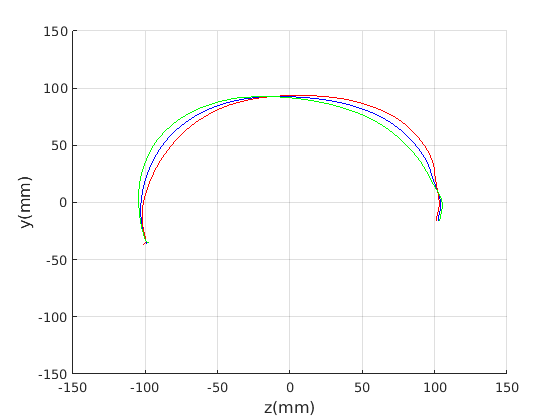} \\
\includegraphics[width=0.4\linewidth]{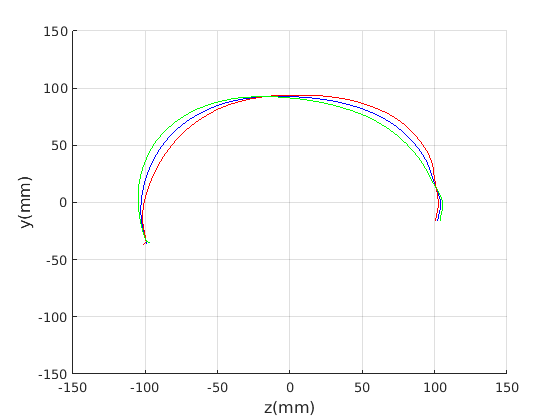} & \includegraphics[width=0.4\linewidth]{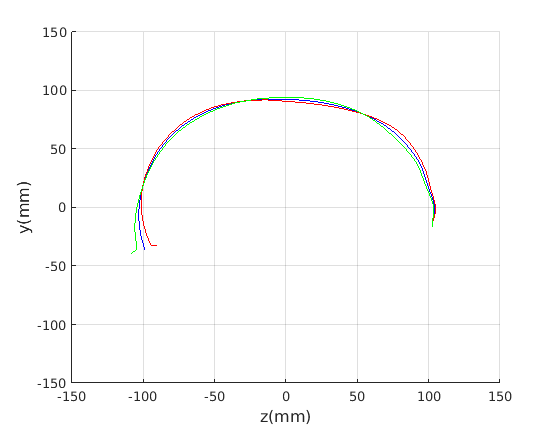} \\
\includegraphics[width=0.4\linewidth]{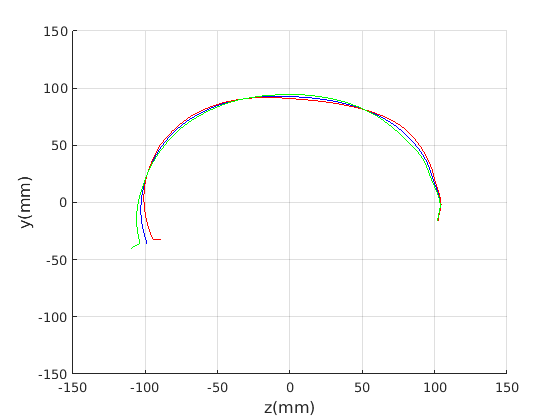} & \includegraphics[width=0.4\linewidth]{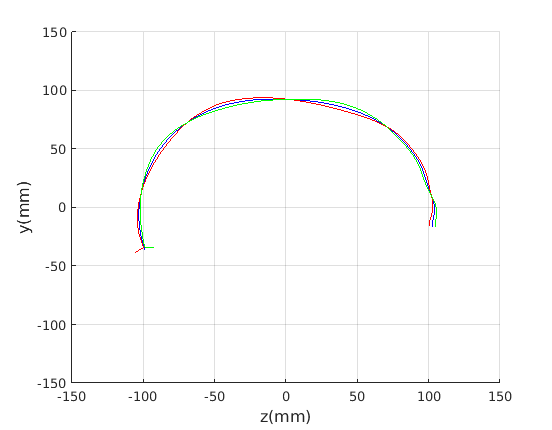} 
\end{tabular}
\caption{Left column: The dominant four modes (Top: mode 1; bottom: mode 4) of cranial shape variation. Mean is blue, mean+3SD is red and mean-3SD is green. Right column: its scale normalised variant.}
\label{fig:cranProfECN}
\end{figure}
\end{center}

To emphasise the form/shape variation for each of these cases, the mean shape and shapes at mean $\pm 3SD$ for the four most significant modes of shape variation are plotted in Fig. \ref{fig:headProfECN} and  Fig. \ref{fig:cranProfECN}. In this case, ECN normalisation is used.

For the craniofacial sagittal profile model, when not scale-normalised (Fig. \ref{fig:headProfECN}, left column), the following four dominant shape modes are observed:

\begin{enumerate}
\item \emph{Cranial height, calvarium convexity and facial angle} constitute the main correlated shape variations captured im mode 1, with small cranial heights being correlated with a depression in the region of the bregma. There appears to be a \emph{waisting} variation where the calvarium alternates between convex (high cranial height) and concave (low cranial height) configurations. Cranial height and 'waisting' are also strongly correlated with the angle of the face relative to the line between the cranial ellipse centre and nasion.

\item The overall size of the head varies : surprisingly this appears to be almost uncorrelated with craniofacial profile shape.
This was only found in the ECN method of pose normalisation.

\item The length of the face varies - there is variation in the ratio of face and cranium size.

\item Variation in the chin size. A smaller chin appears to be correlated with a more prominent forehead.

\end{enumerate}

The fact that the facial orientation appears to rotate initially led us to suspect a flawed pose normalisation procedure.
However, the point about ECN normalisation is that it focuses on aligning the crania rather than faces, using its centre and the nasion, and the face has to follow that normalisation. Thus these results should be interpreted with a fixed cranial ellipse centre and fixed \emph{ellipse centre to nasion} angle in mind.

For the cranial sagittal profile model, when not scale-normalised (Fig. \ref{fig:cranProfECN}, left column), the following  dominant shape modes are observed:

\begin{enumerate}
\item Cranial height variation with a low correlation with cranial length.
\item Cranial length variation with a low correlation with cranial height.
\item A cranial bulge that varies between the front and rear of the cranium.
\item This mode picks up minor variations on cranial shape and some noise where hair exits from the latex cap.
\end{enumerate}

When data is scale normalised (Fig. \ref{fig:cranProfECN}, right column), cranial height and length are more significantly correlated, as shown in the first mode (top right in figure). The second mode model a bulge that can vary from the front to the rear of the cranium. Taken together, these two modes capture close to 90\% of the variation in the training set. A two-dimensional model, based on these two modes is used in a clinical case study in Sect. \ref{sec:caseStudy}.

\begin{center}
\begin{figure}
\begin{tabular}{cc}
\includegraphics[width=0.4\linewidth]{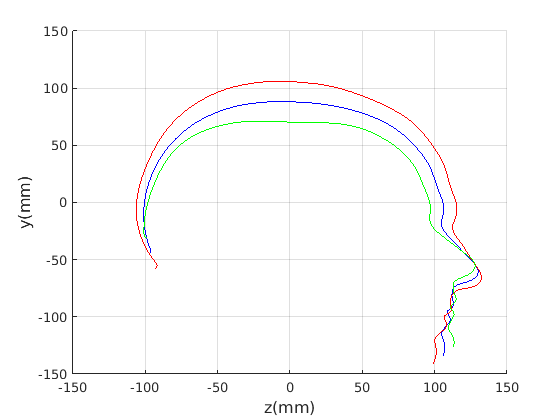} & \includegraphics[width=0.4\linewidth]{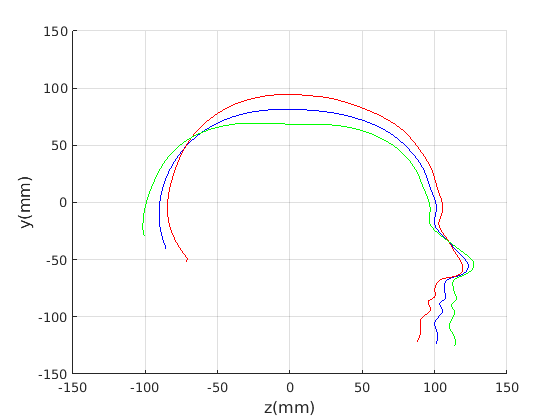} \\
\includegraphics[width=0.4\linewidth]{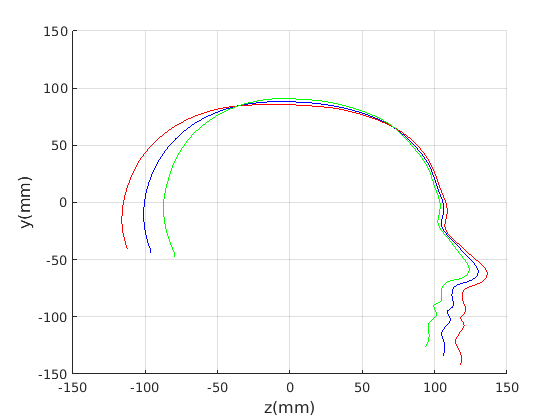} & \includegraphics[width=0.4\linewidth]{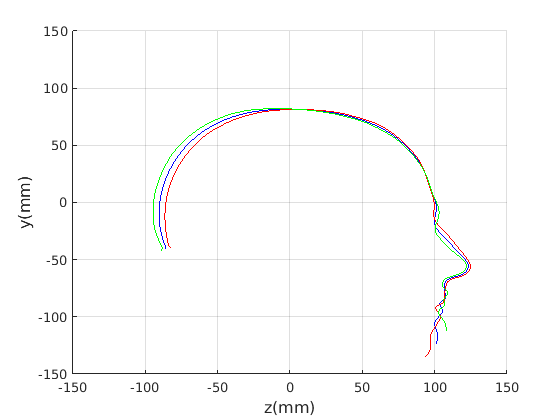} \\
\includegraphics[width=0.4\linewidth]{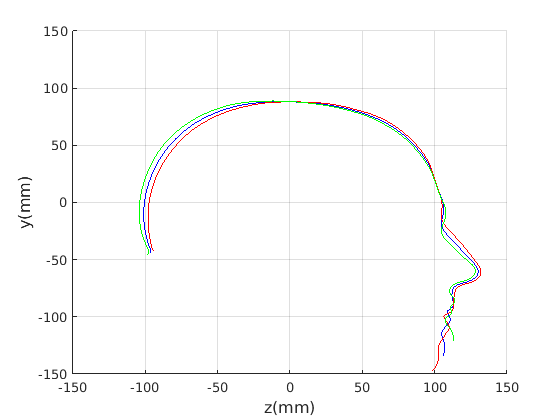} & \includegraphics[width=0.4\linewidth]{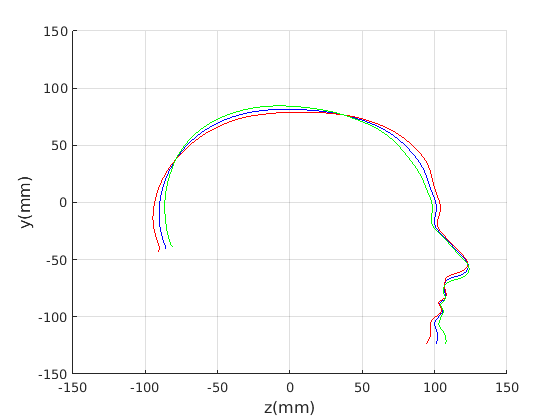} \\
\includegraphics[width=0.4\linewidth]{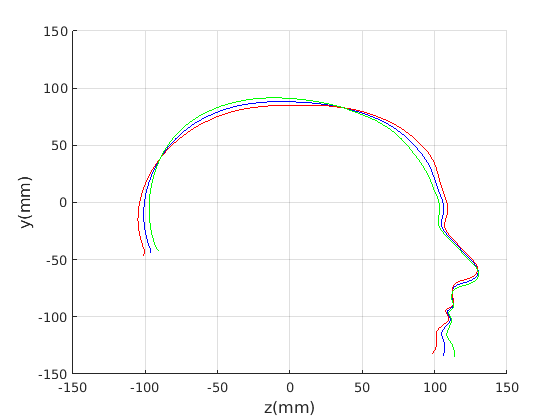} & \includegraphics[width=0.4\linewidth]{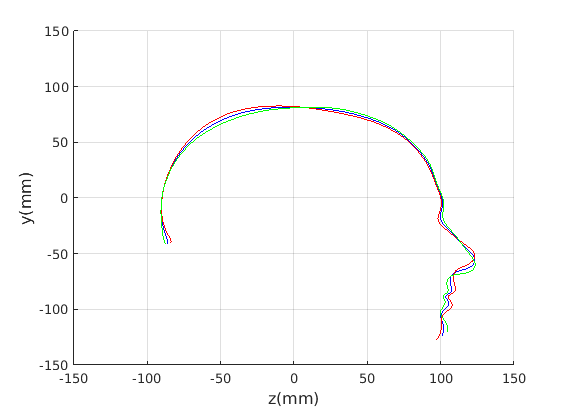} 
\end{tabular}
\caption{Left column: The dominant four modes of craniofacial profile variation in the GPA-aligned model. Right column: its size normalised variant.}
\label{fig:headProfGPA}
\end{figure}
\end{center}

To provide a comparsion with ECN, the form/shape variation modes extracted from a GPA-based alignment are given in 
Fig. \ref{fig:headProfGPA} for full head and Fig. \ref{fig:headProfGPA} for cranium only.

\begin{center}
\begin{figure}
\begin{tabular}{cc}
\includegraphics[width=0.4\linewidth]{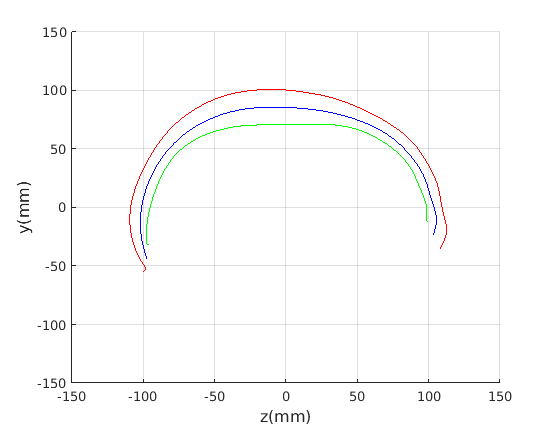} & \includegraphics[width=0.4\linewidth]{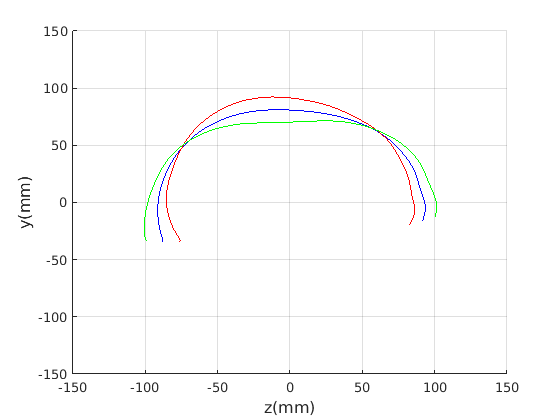} \\
\includegraphics[width=0.4\linewidth]{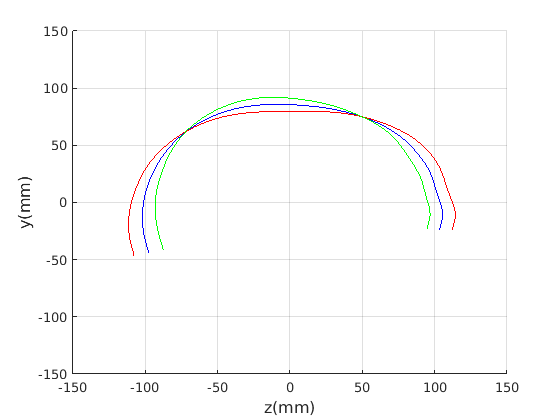} & \includegraphics[width=0.4\linewidth]{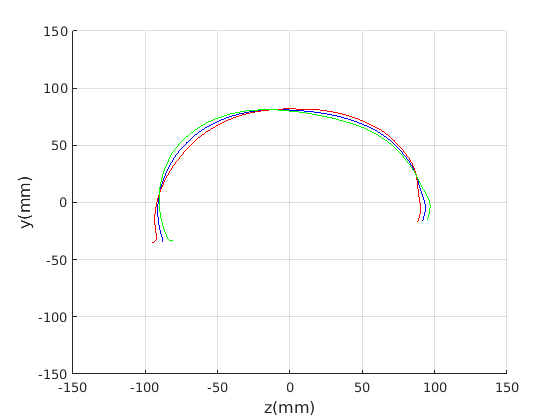} \\
\includegraphics[width=0.4\linewidth]{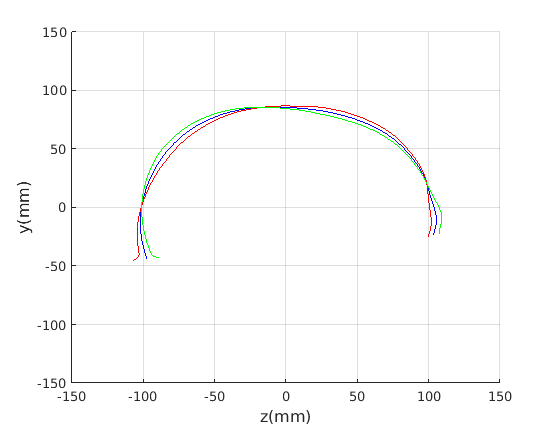} & \includegraphics[width=0.4\linewidth]{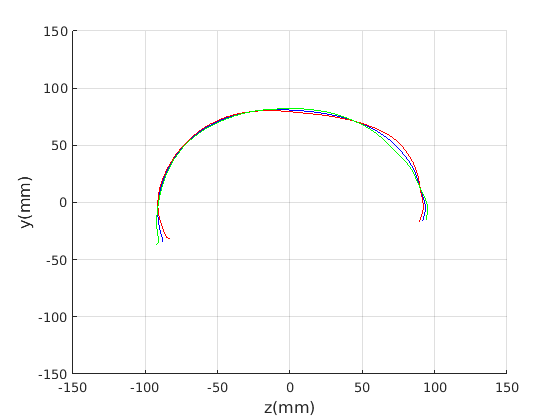} \\
\includegraphics[width=0.4\linewidth]{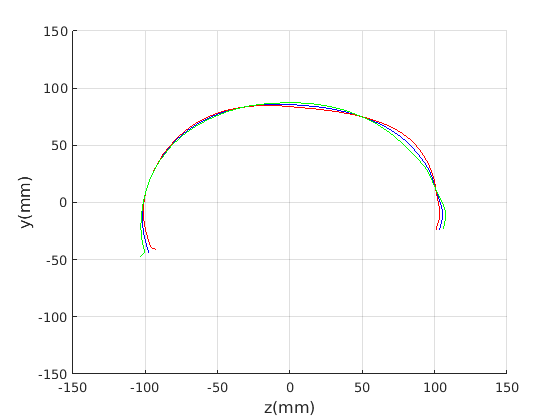} & \includegraphics[width=0.4\linewidth]{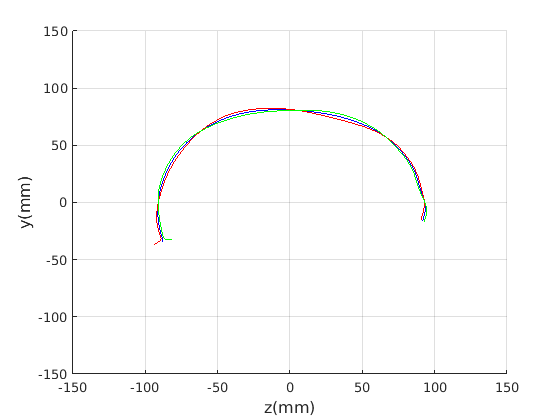} 
\end{tabular}
\caption{Left column: The dominant four modes of cranial shape variation in the GPA-aligned model. Right column: its size normalised variant (right column).}
\label{fig:cranProfGPA}
\end{figure}
\end{center}

\subsection{Model compactness}
\label{sec:compactness}

\begin{figure}
\centerline{\includegraphics[width=0.7\linewidth]{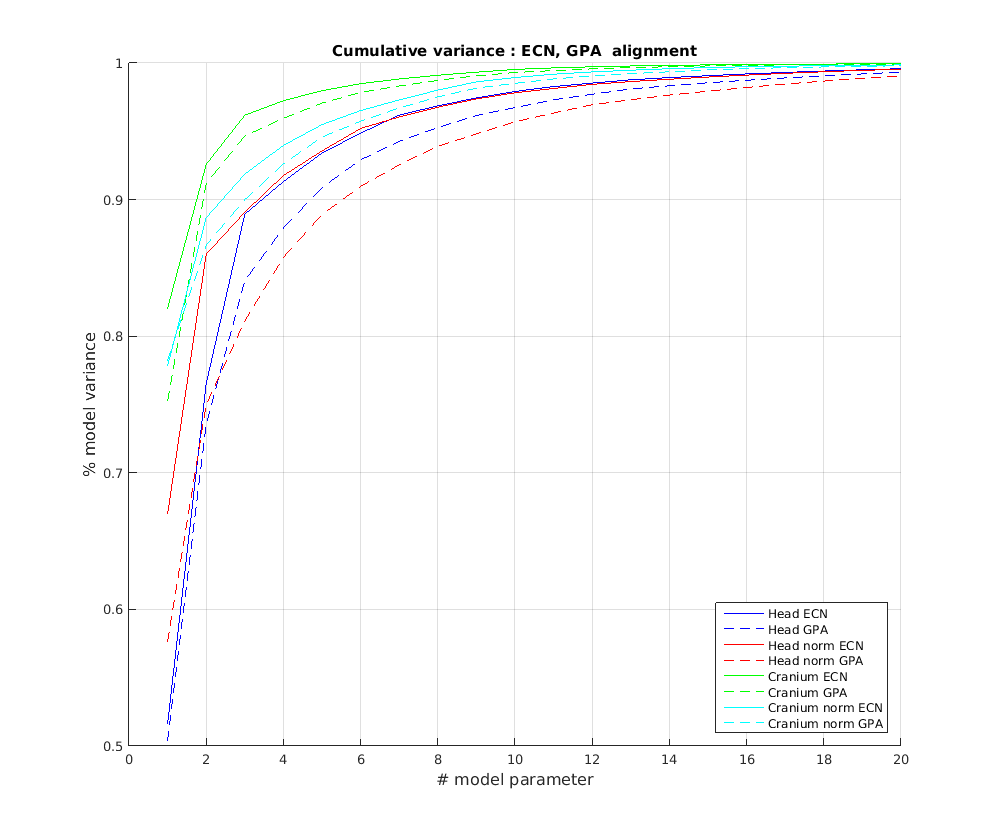} }   
\caption{Cumulative variance plot of head profile models - automatic profile landmarking.}
\label{fig:cumulVarProfECNAut}
\end{figure}

\begin{figure}
\centerline{\includegraphics[width=0.8\linewidth]{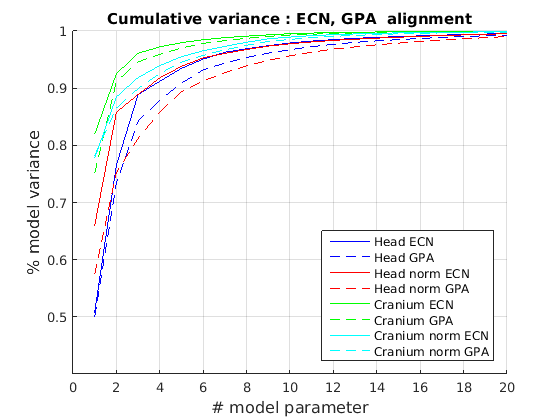} }   
\caption{Cumulative variance plot of head profile models - manual profile landmarking.}
\label{fig:cumulVarProfECNMan}
\end{figure}

We evaluate both model construction with geometric alignment (ECN normalisation) and model construction with alignment by Generalised Procrustes Analysis (GPA). Model compactness is a key performance criterion for the correspondence and alignment processes that generate that model \cite{Styner03}. A more compact model has a smaller variance and requires fewer parameters to express a particular shape instance than less compact model. Cumulative variance plots generated by application of PCA are a useful measure of compactness, with more rapidly rising curves indicating more compact models. 
In Fig. \ref{fig:cumulVarProfECNAut} and Fig. \ref{fig:cumulVarProfECNMan}  we present the cumulative variance plots for our profile models using automatic and manual landmarking respectively and aligned with either ECN or GPA alignment schemes. 

Comparing the compactness of the models produced by automatic and manual facial profile landmarking, we found that the performance is almost identical and there is no statistical significance in performance given the modest test size of 100 faces. For example, for the full head profile, with no scale normalisation, the automatic landmarks give cumulative variances of 51.5\%, 76.5\%, 88.9\% and 91.3\% over the first four modes whereas the manual landmarks give  50.8\%, 76.5\%, 88.9\% and 91.2\%.

A point to note is that whatever model is built, ECN always produces more compact models than GPA. The difference is more marked when a head model is built (cranium and face) compared to cranium only. 
This can be explained by the fact that the cranium has 210 points sampled on its surface, whereas the face has 128.
ECN focusses on aligning the majority group (cranial points) and does not allow more extreme features (such as a large nose) to influence this. GPA on the other hand does, so relatively few points around the nose region can influence alignment over the whole cranium. ECN normalisation is more appropriate when we are interested in cranial shape.

\section{Extension to 3D full head modelling}
\label{sec:3Dmodelling}

Using the ECN pose normalisation technique described in previous sections, we have built a 3D statistical model of the full
head, including face \emph{and} cranium, using the same 100 scans of adult males as we used in the cranial profile model described earlier. We employed a variant of a template morphing technique called Optimal-Step Non-rigid Iterative Closest Points (OSNR ICP) \cite{Amberg07}. However, its application to the full human head is more difficult than with face only, due to the lack of well-defined landmarks for initial alignment over the cranial area.

\begin{figure}
\centerline{\includegraphics[width=0.8\linewidth]{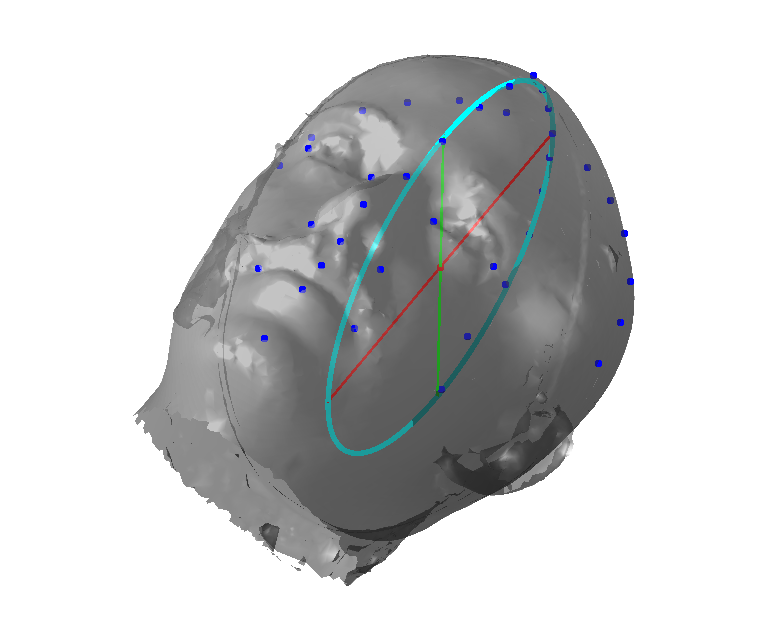}}
\caption{Ellipse fitting on the symmetry plane used to align the scan data and generate cranial pseudo landmarks.}
\label{fig:head3DEllipse}
\end{figure}

\begin{center}
\begin{figure}
\begin{tabular}{cc}
\includegraphics[width=0.5\linewidth]{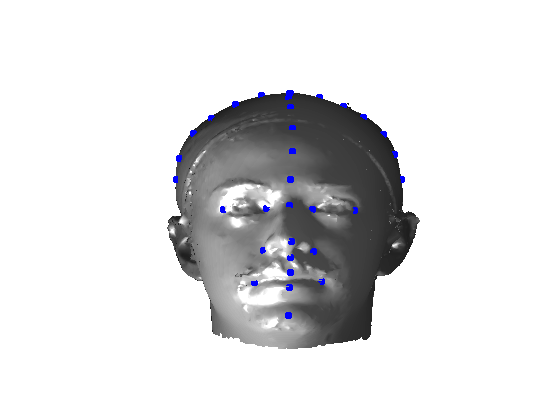} & \includegraphics[width=0.5\linewidth]{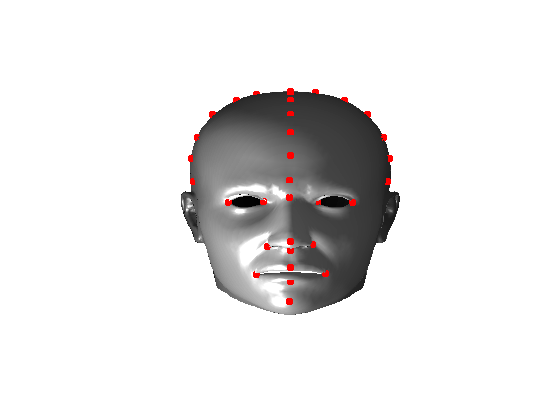} \\
\includegraphics[width=0.5\linewidth]{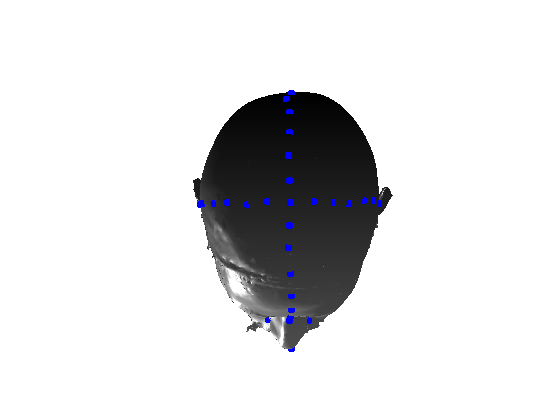} & \includegraphics[width=0.5\linewidth]{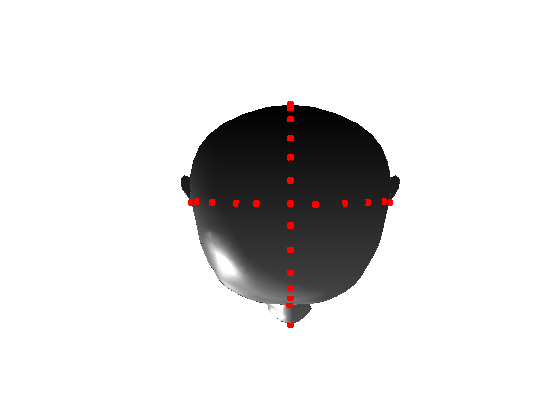} \\
\end{tabular}
\caption{Generating pseudo-landmarks on the cranial surface by projection from the ellipse centre.}
\label{fig:fullHeadModel-pseudoLdmks}
\end{figure}
\end{center}

Pose is normalised using the ECN method, and relies on the symmetry plane, ellipse centre and nasion position, as shown in Fig. \ref{fig:head3DEllipse}.
We then find a set of pseudo-landmarks on the cranial surface by 3D ray projection over a predefined set of angles, as shown in Fig.  \ref{fig:fullHeadModel-pseudoLdmks}. The same processes are applied to a template mesh of the human head, so that it has the same set of both face and cranium landmarks. 

\begin{center}
\begin{figure}
\begin{tabular}{cc}
\includegraphics[width=0.5\linewidth]{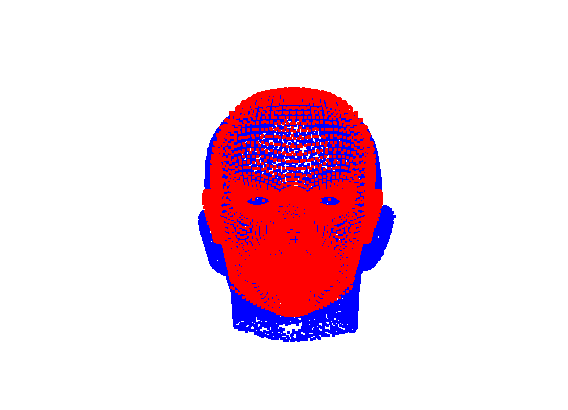} & \includegraphics[width=0.5\linewidth]{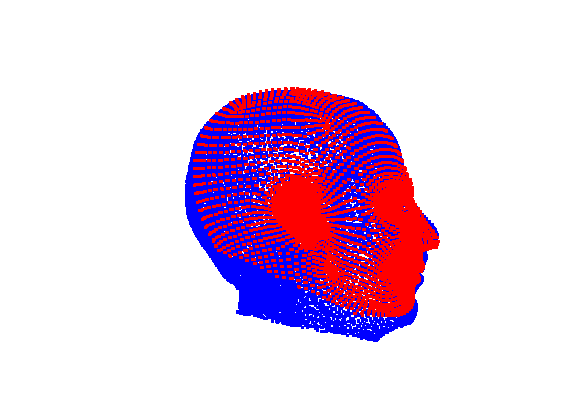} \\
\end{tabular}
\caption{Global affine warping and alignment of the model to the scan using face landmarks and cranium pseudo-landmarks. }
\label{fig:globalAffineWarping}
\end{figure}
\end{center}

The template mesh is then warped in a global affine sense, as shown in Fig. \ref{fig:globalAffineWarping} to minimise the least-squares error between the two sets of landmarks, when the head scan is fixed in its canonical poose. After this initial stage, landmarks and pseudo-landmarks are used in the same way as any other mesh vertex, as they are potentially noisy, as seen in earlier results analysis. This contrasts with Amberg et al's \cite{Amberg07} approach where the influence of landmarks is gradually faded out. 

\begin{center}
\begin{figure}
\begin{tabular}{cc}
\includegraphics[width=0.4\linewidth]{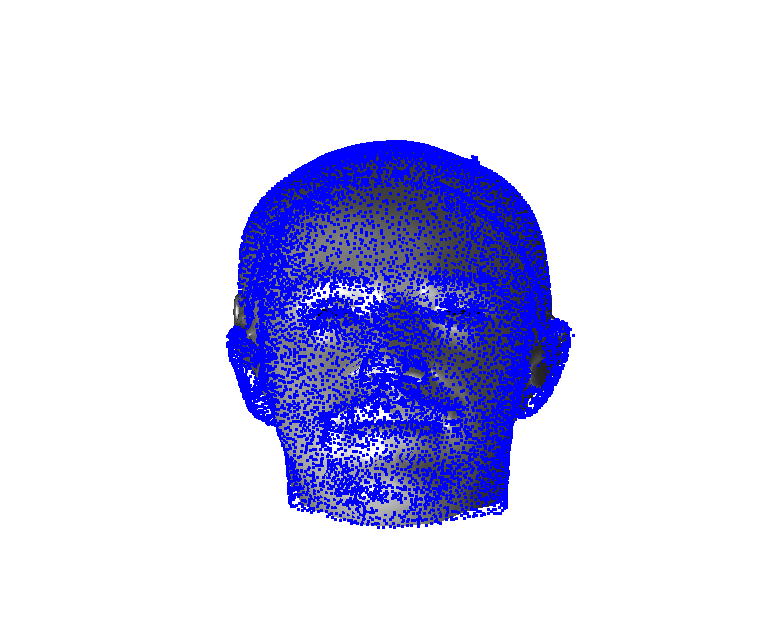} & \includegraphics[width=0.4\linewidth]{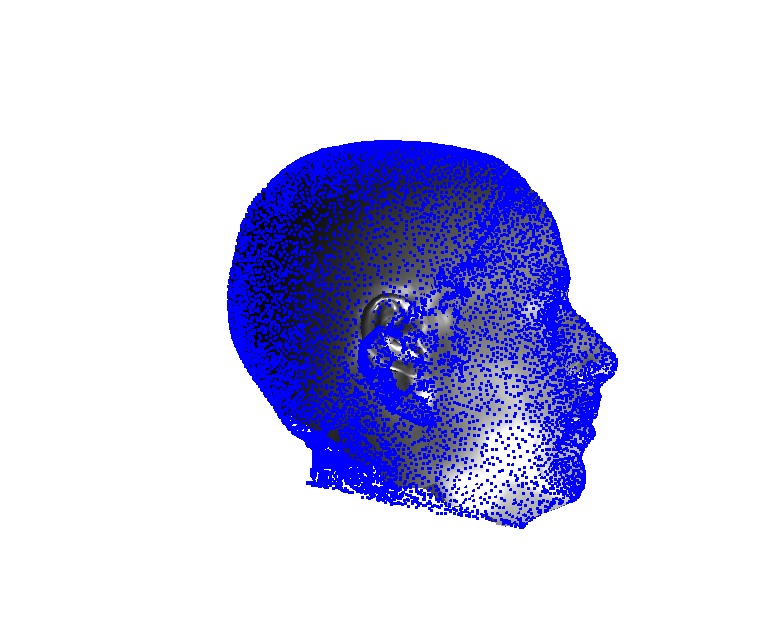} \\
\includegraphics[width=0.4\linewidth]{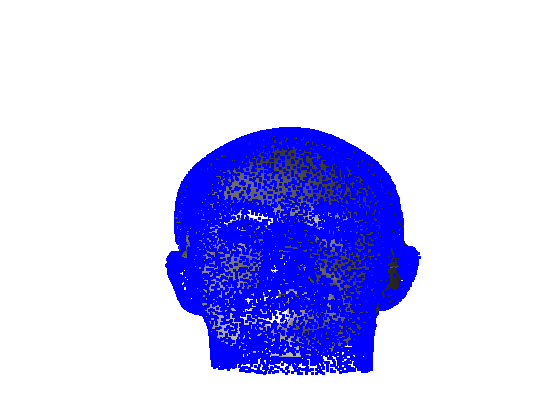} & \includegraphics[width=0.4\linewidth]{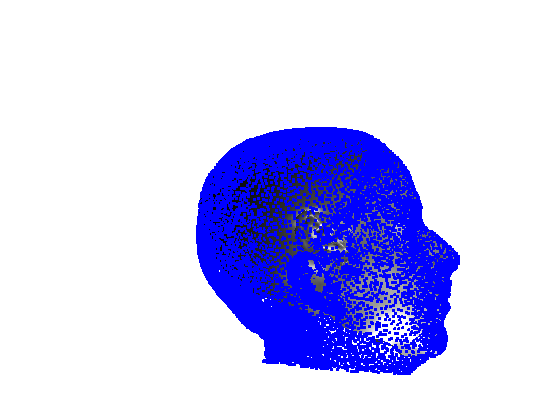} \\
\includegraphics[width=0.4\linewidth]{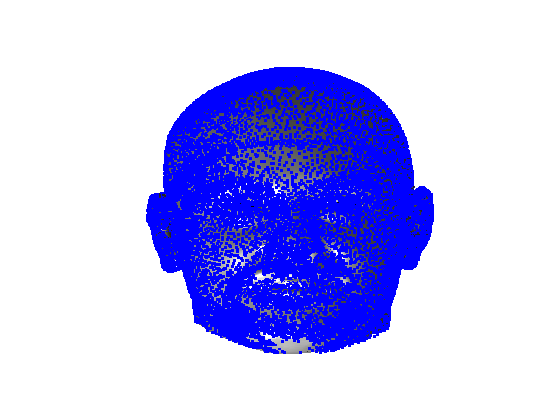} & \includegraphics[width=0.4\linewidth]{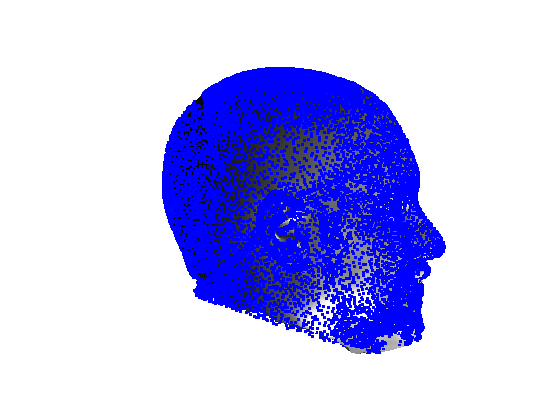} \\
\includegraphics[width=0.4\linewidth]{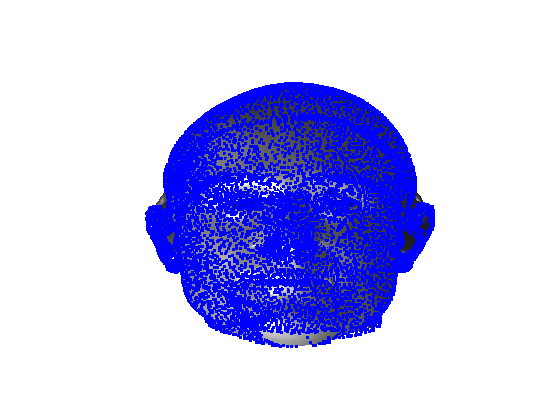} & \includegraphics[width=0.4\linewidth]{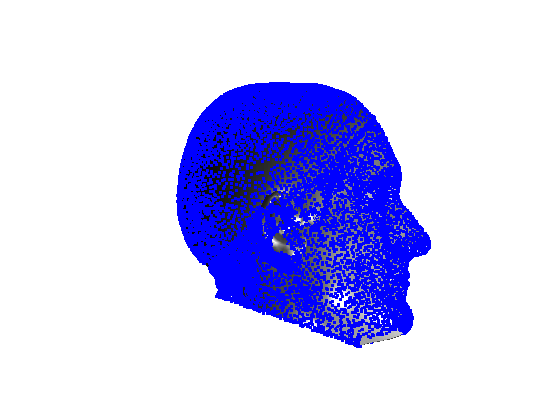} \\
\end{tabular}
\caption{Results of OSNR ICP template warping on the first 4 training scans. Headspace data (downsampled) is shown in blue. Template morph is rendered as a grey surface.}
\label{fig:osnrICPResults}
\end{figure}
\end{center}

Following this, OSNR ICP iterates over a set of decreasing mesh stiffnesses (typically 10), until the registration between the 
template and the scan is below some error or a maximum number of iterations is reached. 
Figure \ref{fig:osnrICPResults} shows a selection of template warp results. 
The template appears to be accurately warped onto the scan, at least in the normal direction of the surfaces, it is not possible to see error tangential to the surface. Furthermore, we have noted errors on some scans on the ears. To improve these we need a method of automatically landmarking the ears, which is an area for further work.

\begin{center}
\begin{figure}
\begin{tabular}{cc}
\includegraphics[width=0.5\linewidth]{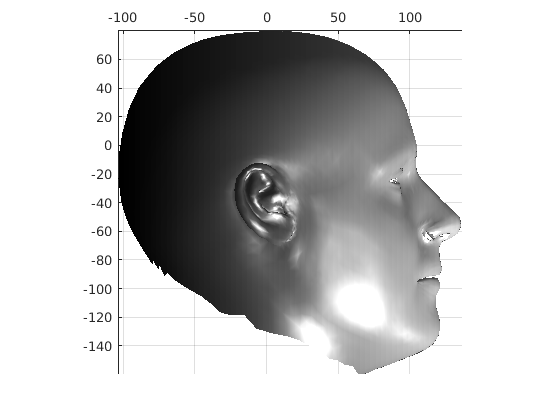} & \includegraphics[width=0.5\linewidth]{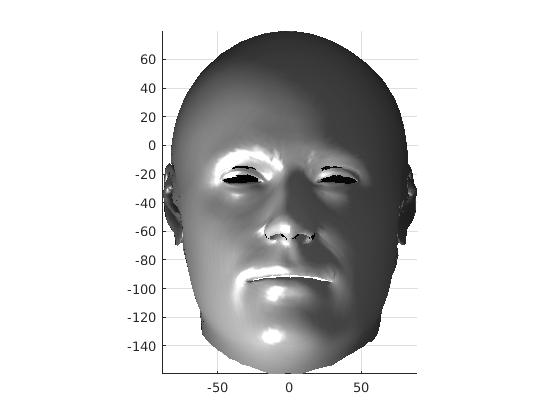} \\
\includegraphics[width=0.5\linewidth]{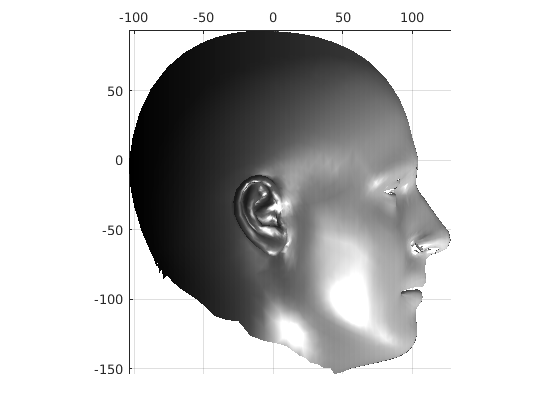} & \includegraphics[width=0.5\linewidth]{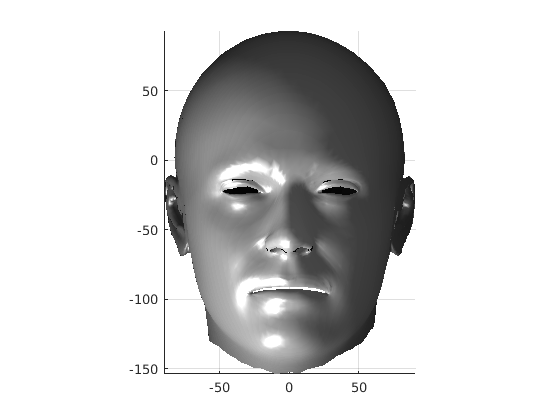} \\
\includegraphics[width=0.5\linewidth]{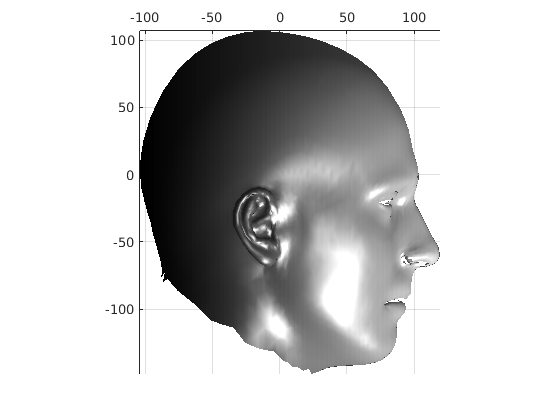} & \includegraphics[width=0.5\linewidth]{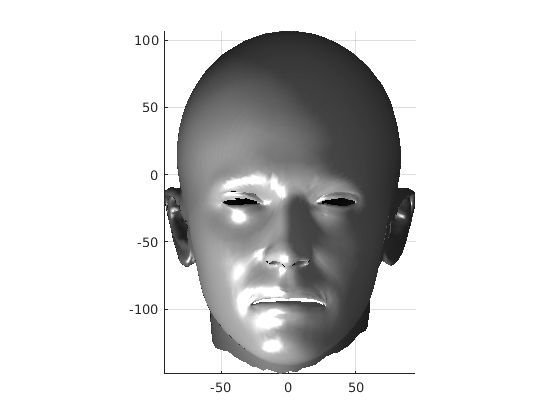} \\
\end{tabular}
\caption{The model of the full head (no scale normalisation, ECN pose normalisation) and its first mode of variation. Note the correlation between cranial height and facial angle. Centre is mean, top is mean -3SD, bottom is mean +3SD.}
\label{fig:fullHeadModel-mode1}
\end{figure}
\end{center}

\begin{center}
\begin{figure}
\begin{tabular}{cc}
\includegraphics[width=0.5\linewidth]{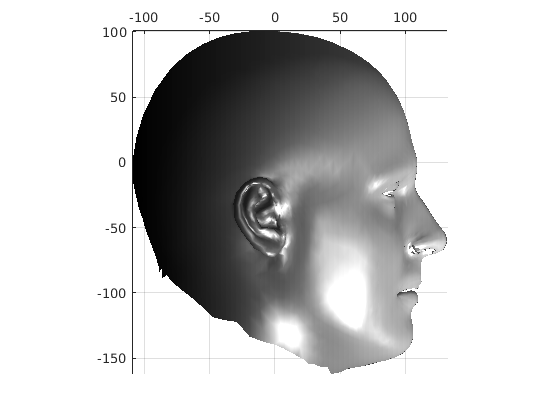} & \includegraphics[width=0.5\linewidth]{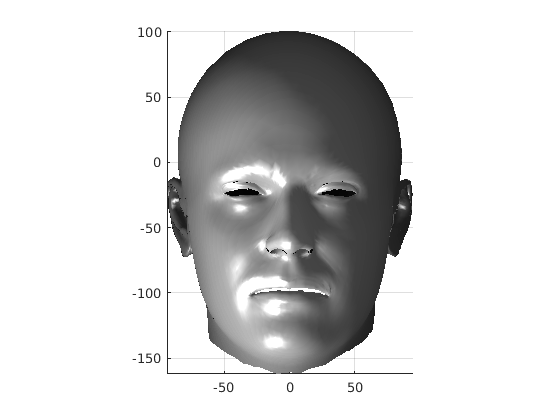} \\
\includegraphics[width=0.5\linewidth]{fhmMeanSide.png} & \includegraphics[width=0.5\linewidth]{fhmMeanFront.png} \\
\includegraphics[width=0.5\linewidth]{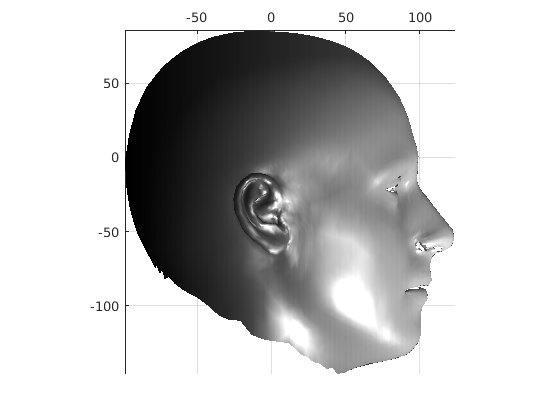} & \includegraphics[width=0.5\linewidth]{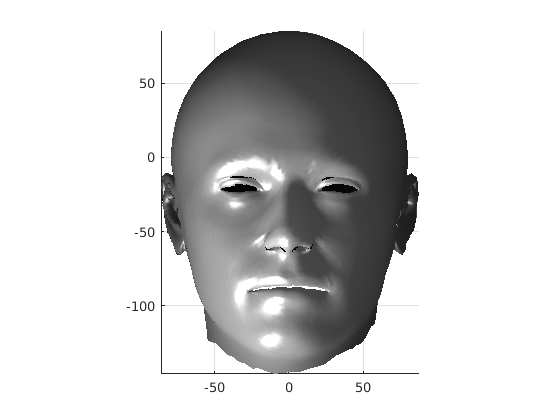} \\
\end{tabular}
\caption{The model of the full head (no scale normalisation, ECN pose normalisation) and its second mode of variation. This is an almost pure size variation. Centre is mean, top is mean -3SD, bottom is mean +3SD.}
\label{fig:fullHeadModel-mode2}
\end{figure}
\end{center}

The first two modes of head variation are shown in figures \ref{fig:fullHeadModel-mode1} and \ref{fig:fullHeadModel-mode2}.
These include the cranial height / facial angle mode and the (almost) pure size mode, as seen in the sagittal profile model.

\section{A case study of cranial profile model use for intervention outcome evaluation}
\label{sec:caseStudy}

\begin{figure}
\centerline{\includegraphics[width=0.8\linewidth]{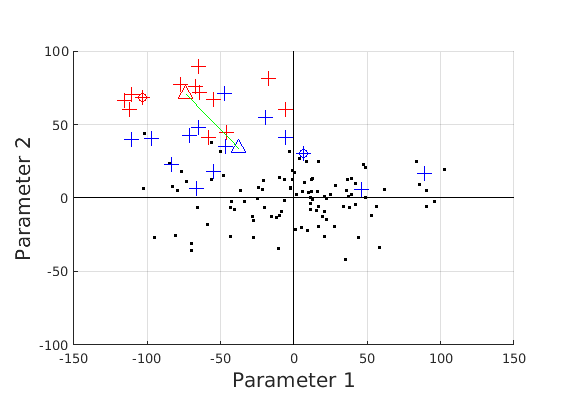} }   
\caption{Patient cranial profile parameterisations, BS intervention: pre-operative (red crosses) and post-operative (blue crosses) in comparison to the training set of 100 scans (black dots) : mean cranial shape is at the origin. The circled values represent an example patient, presented in Fig \ref{fig:BSexample}. }
\label{fig:bsResults}
\end{figure}

\begin{figure}
\centerline{\includegraphics[width=0.8\linewidth]{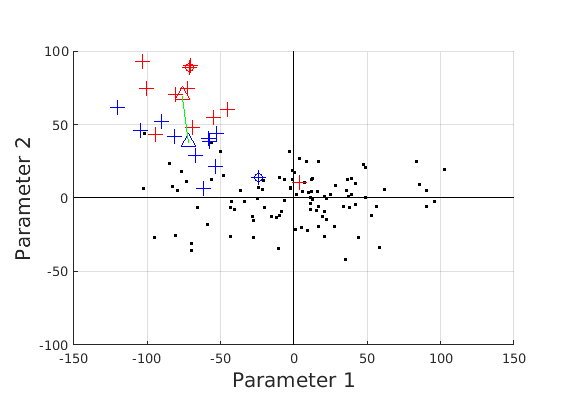} }   
\caption{Patient cranial profile parameterisations, TCR intervention: pre-operative (red crosses) and post-operative (blue crosses) in comparison to the training set of 100 scans (black dots) : mean cranial shape is at the origin. The circled values represent an example patient, presented in Fig \ref{fig:TCRexample}.}
\label{fig:tcrResults}
\end{figure}

In this section we take a sample of 25 boys, who are craniofacial craniosynostosis patients, 
14 of which have undergone one type of corrective procedure (BS) and the other 11, another corrective procedure (TCR). 
Providing that the heads are sufficiently symmetrical and are amenable to ellipse fitting, 
we can parameterise all of these patients' heads using our scale-normalised craniofacial profile model (2D model with face cropped out).
We can then plot their pre-operative and post-operative parametrisations and compare them with the parameterisations of the 
100 training examples. The expected result is that the parameterisations should show the head shapes moving nearer to the 
mean of the training examples. It also reveals which of the dominant modes of shape variation are most affected.
The results are shown in figures \ref{fig:bsResults} and \ref{fig:tcrResults}.

For the BS patient set, the Mahalanobis distance of the mean pre-op parameters (red triangle in Fig. \ref{fig:bsResults}) is 4.670, and for the mean post-op parameters (blue triangle) is 2.302. For shape parameter 2 only these figures are 4.400 and 2.156.

For the TCR patient set, the Mahalanobis distance of the mean pre-op parameters (red triangle in Fig. \ref{fig:tcrResults}) is 4.647, and for the mean post-op parameters (blue triangle) is 2.439. For shape parameter 2 only these figures are 4.354 and 2.439.
We note that most of this change occurs in parameter 2, which corresponds to moving height in the cranium from the frontal part of the profile to the rear. In these figures we excluded one patient, who preoperatively already had a near-mean head shape (see red cross near to the origin in Fig. \ref{fig:tcrResults}, so any operation is unlikely to improve on this (but intervention is required in order to relieve potentially damaging inter-cranial pressure).

\begin{center}
\begin{figure}
\begin{tabular}{cc}
\includegraphics[width=0.5\linewidth]{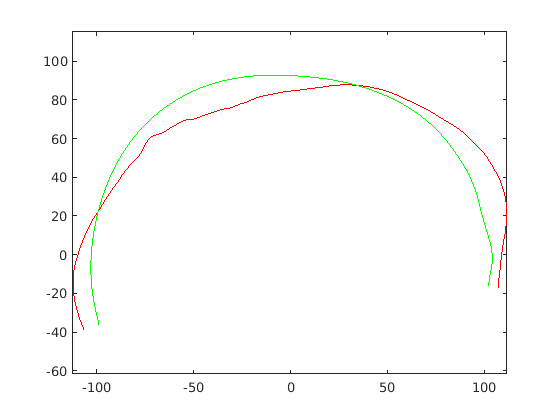} & \includegraphics[width=0.5\linewidth]{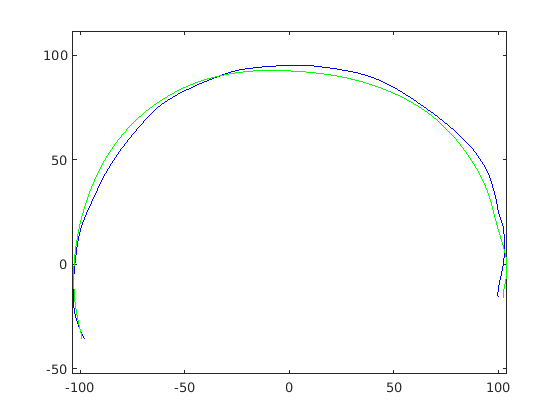} \\
\end{tabular}
\caption{Pre-op and post-op profiles for a BS patient. The red and blue traces show the extracted sagittal profiles (scale normalised) of the patient pre-operatively and post-operatively respectively, whilst the green shows the mean profile of the training set.}
\label{fig:BSexample}
\end{figure}
\end{center}

\begin{center}
\begin{figure}
\begin{tabular}{cc}
\includegraphics[width=0.5\linewidth]{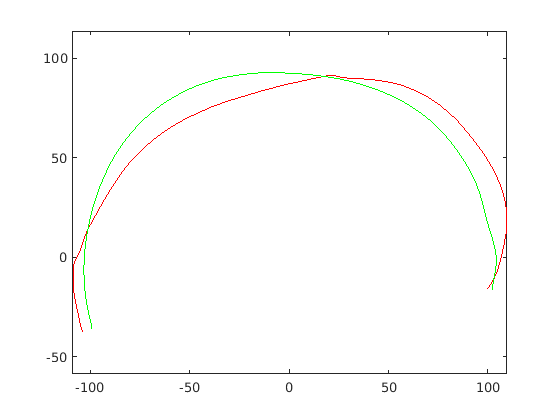} & \includegraphics[width=0.5\linewidth]{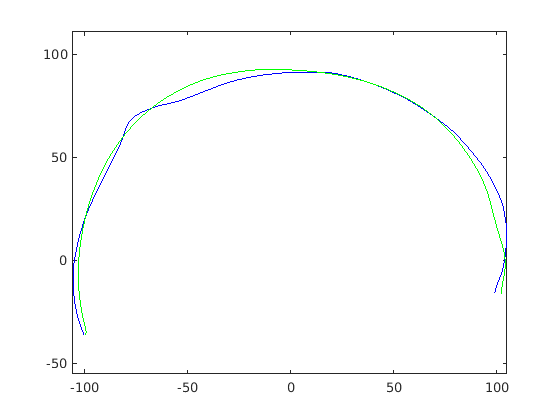} \\
\end{tabular}
\caption{Preop and postop profiles for a TCR patient. The red and blue traces show the extracted sagittal profiles (scale normalised) of the patient pre-operatively and post-operatively respectively, whilst the green shows the mean profile of the training set.}
\label{fig:TCRexample}
\end{figure}
\end{center}

It is not possible to make definitive statements relating to one method of intervention compared to another with these relatively small numbers of patients. However, the cranial profile model does show that both procedures on average, lead to a movement of head shape towards the mean of the training population. An example of analysis of intervention outcome for a  BS patient is given in Fig. \ref{fig:BSexample} and a TCR patient is given in Fig. \ref{fig:TCRexample}. The particular example used is highlighted with circles on figures \ref{fig:bsResults} and \ref{fig:tcrResults} to indicate pre-op and post-op parametrisations. To our knowledge this is the first use of statistical 3D craniofacial shape models in a clinical study.

\section{Conclusions and further work}
\label{sec:conclusions}

We have built a statistical model of craniofacial profile shape variation, and used this to initialise a template morphing algorithm for a full 3D model of the human head. The various shape models have revealed a number of interesting modes of head shape variation among adult males. 

Cranial height, calvarium convexity and facial angle constitute the main correlated shape variations (captured im mode 1), with small cranial heights being correlated with a depression in the region of the bregma, which we term a `waisting' variation.
Other major modes include variation in the ratio of cranium and face size and a cranial bulge that varies from the front to the rear of the cranium.

The automatic model building process employed first uses a machine learning technique to extract landmarks, then various ICP stages to extract global and local symmetry planes and hence the head profile by intersection with the mesh. Cubic splines are used to smooth noise on the profile, fill holes, present a mechanism for landmark position refinement, and a means of geodesic interpolation to generate profile model points. Ellipse fitting along with nasion localiation allowed a novel pose normalisation technique which yielded are more compact model than standard GPA alignment. The model's use in craniofacial intervention outcome assessment was demonstrated.

Further work will expand our model building to the full headspace dataset. We will build models for various demographic partitions and explore the performance of a wider variety of model building techniques and make the \emph{Headspace} dataset public for other researchers to compare their results with ours.

\section*{References}

\bibliography{mia}

\end{document}